\documentclass{article}

\usepackage{arxiv}

\usepackage[utf8]{inputenc} % allow utf-8 input
\usepackage[T1]{fontenc}    % use 8-bit T1 fonts
\usepackage{hyperref}       % hyperlinks
\usepackage{url}            % simple URL typesetting
\usepackage{booktabs}       % professional-quality tables
\usepackage{amsfonts}       % blackboard math symbols
\usepackage{nicefrac}       % compact symbols for 1/2, etc.
\usepackage{microtype}      % microtypography
\usepackage{cleveref}       % smart cross-referencing
\usepackage{graphicx}
\usepackage{natbib}
\usepackage{doi}
\usepackage{array}
\usepackage{multirow}
\usepackage{longtable}
\usepackage{rotating}

\title{ConvoMem Benchmark: Why Your First 150 Conversations Don't Need RAG}

\date{}

\newif\ifuniqueAffiliation
\uniqueAffiliationtrue

\author{
    Egor Pakhomov\\
    Salesforce AI Research\\
    \texttt{egor.pakhomov@salesforce.com}
    \And
    Erik Nijkamp\\
    Salesforce AI Research\\
    \texttt{erik.nijkamp@salesforce.com}
    \And
    Caiming Xiong\\
    Salesforce AI Research\\
    \texttt{cxiong@salesforce.com}
}

\renewcommand{\shorttitle}{ConvoMem Benchmark}

\hypersetup{
pdftitle={ConvoMem Benchmark: Why Your First 150 Conversations Don't Need RAG},
pdfsubject={cs.CL, cs.AI, cs.IR},
pdfauthor={Egor Pakhomov, Erik Nijkamp, Caiming Xiong},
pdfkeywords={conversational memory, retrieval-augmented generation, RAG, benchmarking, long context, memory systems},
}

\begin{document}
\maketitle

\begin{abstract}
We introduce a comprehensive benchmark for conversational memory evaluation containing
75,336 question-answer pairs across diverse categories including user facts, assistant
recall, abstention, preferences, temporal changes, and implicit connections. While existing benchmarks like LongMemEval and LoCoMo have
advanced the field, our work addresses fundamental challenges in statistical power, data
generation consistency, and evaluation flexibility that limit current memory evaluation
frameworks. We examine the relationship between
conversational memory and retrieval-augmented generation (RAG). While these systems share
fundamental architectural patterns---temporal reasoning, implicit extraction, knowledge updates,
and graph representations---memory systems have a unique characteristic: they start from zero
and grow progressively with each conversation. This characteristic enables naive
approaches that would be impractical for traditional RAG. Consistent with recent findings
on long context effectiveness, we observe that simple full-context approaches achieve
70-82\% accuracy even on our most challenging multi-message evidence cases, while sophisticated
RAG-based memory systems like Mem0 achieve only 30-45\% when operating on conversation
histories under 150 interactions. Our analysis reveals
practical transition points: long context excels for the first 30 conversations, remains
viable with manageable trade-offs up to 150 conversations, and typically requires hybrid
or RAG approaches beyond that point as costs and latencies become prohibitive. We also
find that medium-tier models deliver equivalent memory performance to
premium models at 8x lower cost, suggesting significant optimization opportunities. These
patterns indicate that the small-corpus advantage of conversational memory---where exhaustive
search and complete reranking are feasible---deserves dedicated research attention rather
than simply applying general RAG solutions to conversation histories.
\end{abstract}

\section{Introduction}
\label{sec:introduction}

For the purposes of this paper, we use ``conversational memory'' as a convenient umbrella
term for a chat assistant's ability to leverage prior conversations to provide more
helpful and contextually aware responses. While the literature already addresses
semantic, episodic, and procedural memory as distinct components, we find it useful to
operate at this higher level of abstraction when discussing systems that combine these
various memory types in practice. This framing helps us analyze the overall challenge of
making dialogue systems more natural and effective through memory, without getting caught
up in the boundaries between different memory categories.

When we examine conversational memory as a research area, we find it has substantial
overlap with retrieval-augmented generation (RAG). RAG is a research area focused on
systems that enhance language model responses by retrieving relevant information from a
pre-existing corpus, then incorporating that retrieved context into the generation process. Both memory and RAG can be fundamentally
characterized as making an agent more helpful by leveraging context captured in some text
corpus. In memory systems, this corpus consists of prior conversations; in RAG systems,
it comprises documents, web content, or enterprise text accumulations. A perfect system
that knows how to incorporate information from its corpus into every response would excel
at both memory and RAG tasks. This raises a critical question: are the most fruitful
research directions in memory those that significantly overlap with RAG, or should we
focus on memory-specific optimizations?

To address this question, we need both better evaluation tools and empirical evidence about what makes memory systems unique. This paper makes two primary contributions:
\begin{itemize}
\item We introduce a large-scale, statistically robust benchmark for evaluating conversational
  memory systems with 75,336 question-answer pairs across diverse categories---including
  those from LongMemEval as well as additional categories we developed---with design emphasis
  on enterprise scenarios like customer relationship management, technical support, and
  business process automation, ensuring comprehensive coverage of memory capabilities
\item We analyze the fundamental relationship between memory and RAG systems, demonstrating
  that memory systems can leverage unique characteristics---particularly their smaller,
  progressively growing search spaces---to achieve strong performance with naive policies
\end{itemize}

\section{A Comprehensive Benchmark for Conversational Memory}
\label{sec:benchmark}

\subsection{Limitations of Existing Benchmarks}

The evolution of conversational memory evaluation reveals a field progressing from simple
consistency checks to complex multi-dimensional assessments, yet fundamental limitations
persist across all existing frameworks. Early benchmarks established foundational concepts:
MSC \citep{arxiv:2107.07567} pioneered multi-session evaluation with 5,000 dialogues spanning
up to 5 sessions, while DuLeMon \citep{arxiv:2203.05797} introduced mutual persona memory for 27,000 Chinese
conversations, demonstrating the need for bidirectional memory where agents track both user
and system information. These early efforts, though groundbreaking, operated at scales
(1,000-5,000 tokens) that modern 100K-context models handle trivially, and focused
narrowly on persona consistency rather than comprehensive memory capabilities.

The field underwent significant expansion in 2024 with benchmarks attempting to address
different dimensions of the memory challenge. LongMemEval \citep{arxiv:2410.10813} represents the
current state-of-the-art, introducing valuable conceptual categories for memory evaluation
including user-specific, abstinence, and temporal questions. LoCoMo \citep{arxiv:2402.17753} pushed boundaries with
extremely long conversations---10 synthetic dialogues each spanning 35 sessions and approximately
9,000 tokens---and introduced multimodal elements through image sharing tasks. However, a critical
finding emerged: simple filesystem operations achieved 74\% accuracy on LoCoMo, matching or
exceeding sophisticated memory systems \citep{url:https://www.letta.com/blog/benchmarking-memory}, suggesting the benchmark fails to meaningfully
differentiate between trivial and intelligent memory approaches. PerLTQA \citep{arxiv:2402.16288} took an orthogonal approach with 8,593 questions across 30
fictional characters, explicitly distinguishing between semantic memory (static facts) and
episodic memory (experiences), yet lacks the interactive dialogue dynamics essential for
conversational evaluation. DialSim \citep{arxiv:2406.13144} introduced real-time
evaluation constraints, testing memory retrieval under time pressure across 350,000-token
TV show transcripts, but its entertainment domain specificity and scripted nature limit
generalizability to authentic user-assistant interactions.

Most recently, MemoryAgentBench \citep{arxiv:2507.05257} attempted comprehensive evaluation
through four core competencies---accurate retrieval, test-time learning, long-range
understanding, and conflict resolution---finding that no current system excels across all
dimensions. Yet like all existing benchmarks, this framework operates on insufficient
scale for statistical significance and offers limited coverage of critical capabilities
such as implicit reasoning requiring unstated connections. While it addresses many
dimensions of memory, gaps remain in areas like assistant-side memory tracking and
multi-message evidence synthesis. Concurrently, ImplexConv \citep{arxiv:2503.07018} made
important contributions specifically to implicit reasoning evaluation, introducing 2,500
examples testing opposed and supportive reasoning scenarios where relevant information is
embedded in semantically distant connections. While their hierarchical TaciTree framework
advances retrieval of implicit knowledge, ImplexConv focuses narrowly on this single
aspect of memory rather than providing comprehensive coverage across all memory capabilities.

As the current state-of-the-art, LongMemEval warrants particular scrutiny. Despite its
valuable contributions to categorizing memory evaluation, it suffers from severe statistical
limitations. With only 500 total questions distributed across multiple categories, individual
category sizes become prohibitively small. For instance, the Preferences category contains
merely 30 questions, yielding approximately $\pm$18\% margin of error at 95\% confidence. Some
multi-session subgroups contain as few as 6 questions, resulting in error margins approaching
$\pm$40\%. These sample sizes are insufficient for drawing statistically significant conclusions
or detecting meaningful differences between systems.

Beyond statistical issues, LongMemEval has a critical methodological flaw: it sources filler
conversations from other benchmarks rather than generating them within the same framework as
evidence-containing conversations. This creates a stylistic mismatch that potentially allows
systems to identify evidence conversations based on their different characteristics rather
than through genuine memory retrieval. Such heterogeneity undermines the benchmark's validity,
as models might achieve high scores by exploiting stylistic patterns rather than demonstrating
actual memory capabilities.

Across the entire landscape, benchmarks exhibit consistent methodological flaws: sample sizes per
category often fall below 100 questions (LoCoMo's 10 conversations, LongMemEval's 30
preference questions), making statistical comparisons unreliable; synthetic data generation
introduces detectable patterns that models exploit rather than demonstrating genuine memory;
and the absence of multi-message evidence synthesis means benchmarks test shallow retrieval
rather than complex integration. The disconnect between sophisticated memory architectures
being developed---Mem0's graph-augmented memory achieving 26\% accuracy improvements \citep{arxiv:2504.19413}, Zep's temporal knowledge graphs reducing latency by 90\% \citep{arxiv:2501.13956}, HippoRAG's neurobiologically-inspired retrieval \citep{arxiv:2405.14831}---and the
limited evaluation frameworks available to assess them creates a fundamental bottleneck in
advancing the field toward genuinely capable conversational memory systems.

\begin{table}[h]
\caption{Comparison of Existing Conversational Memory Benchmarks}
\centering
\small
\setlength{\tabcolsep}{4pt}
\begin{tabular}{@{}lcccccccccc@{}}
\toprule
\multirow{2}{*}{Benchmark} & \multirow{2}{*}{Year} & \multirow{2}{*}{\parbox{1.2cm}{\centering Total\\Q/A}} & \multirow{2}{*}{\parbox{1.5cm}{\centering Context\\(tokens)}} & \multicolumn{6}{c}{Evidence Categories} & \multirow{2}{*}{\parbox{1.3cm}{\centering Multi-\\Message$^1$}} \\
\cmidrule(lr){5-10}
 & & & & \rotatebox{90}{User Facts} & \rotatebox{90}{Assistant Facts} & \rotatebox{90}{Abstention} & \rotatebox{90}{Preferences} & \rotatebox{90}{Changing Facts} & \rotatebox{90}{Implicit Conn.} & \\
\midrule
MSC & 2022 & N/A$^2$ & 1k-5k & Partial & $\times$ & $\times$ & $\times$ & $\times$ & $\times$ & Partial \\
DuLeMon & 2022 & N/A$^2$ & $\sim$1k & Partial$^3$ & Partial$^3$ & $\times$ & $\times$ & $\times$ & $\times$ & $\times$ \\
MemoryBank & 2024 & 194 & 5k & $\checkmark$ & $\times$ & $\times$ & $\times$ & $\checkmark$ & $\times$ & $\times$ \\
PerLTQA & 2024 & 8,593 & $\sim$1M$^4$ & $\checkmark$ & $\times$ & $\checkmark$ & $\times$ & $\times$ & $\times$ & $\times$ \\
LoCoMo & 2024 & 7,512 & $\sim$9k & $\checkmark$ & $\times$ & $\checkmark$ & $\times$ & $\checkmark$ & $\times$ & $\checkmark$ \\
LongMemEval & 2024 & 500 & 115k-1.5M & $\checkmark$ & $\checkmark$$^5$ & $\checkmark$ & $\checkmark$ & $\checkmark$ & $\times$ & Partial \\
MemoryAgentBench & 2024 & Hard to map & Varies & $\checkmark$ & $\times$ & $\times$ & $\times$ & $\checkmark$ & Partial & $\checkmark$ \\
DialSim & 2024 & $\sim$1M$^6$ & 350k & $\checkmark$ & $\times$ & $\checkmark$ & $\times$ & $\checkmark$ & $\times$ & Limited$^7$ \\
ImplexConv & 2025 & 2,500 & 60k & N/A & $\times$ & $\times$ & $\times$ & $\times$ & $\checkmark$ & $\times$ \\
\textbf{ConvoMem (Ours)} & 2025 & \textbf{75,336} & \textbf{1k-3M$^8$} & \textbf{$\checkmark$} & \textbf{$\checkmark$} & \textbf{$\checkmark$} & \textbf{$\checkmark$} & \textbf{$\checkmark$} & \textbf{$\checkmark$} & \textbf{$\checkmark$} \\
\bottomrule
\end{tabular}
\label{tab:benchmark_comparison}
\end{table}

\footnotesize{
$^1$ Tests information distributed across multiple messages/sessions \\
$^2$ Dialogue-based evaluation, not Q/A pairs \\
$^3$ DuLeMon tests mutual persona memory implicitly through dialogue consistency, not explicit Q/A \\
$^4$ Estimated from paper \\
$^5$ LongMemEval tests assistant recall in limited scenarios \\
$^6$ Generated questions, not curated \\
$^7$ At most 2 sessions per their paper \\
$^8$ Configurable from 2-300 conversations
}

\normalsize

This comparison reveals several critical gaps in existing benchmarks: (1) statistical power---our 75,336 questions provide 150x more data than LongMemEval's 500, enabling meaningful statistical analysis; (2) comprehensive coverage---we are the only benchmark systematically evaluating all six critical memory capabilities; (3) multi-message evidence---our framework uniquely implements systematic distribution of evidence across 1-6 messages as a core dimension rather than special cases; and (4) methodological consistency---unlike LongMemEval which sources filler from other benchmarks, our unified generation pipeline prevents style-based shortcuts.

\subsection{Question Categories}

Our benchmark evaluates six distinct categories, each testing different capabilities
required for effective conversational memory. These categories progress from
foundational recall tasks to sophisticated reasoning involving assistant response recall,
uncertainty handling, and inferential thinking.

\subsubsection{Multi-Message Evidence Distribution}
Unlike existing benchmarks where information is typically concentrated in single messages,
our framework implements configurable multi-message evidence distribution as a core dimension
across all categories. While other benchmarks treat multi-message scenarios as special cases
or separate categories, we recognize this as a fundamental dimension that applies universally.

For each category, we systematically vary the number of messages across which relevant
information spreads---from single messages to scenarios requiring synthesis across six or
more exchanges. This design reflects real-world conversation patterns where information
naturally distributes across multiple exchanges rather than being stated all at once. A user
might mention their budget in one message, their performance requirements in another, and
their timeline in a third---all days apart. Effective memory systems must gather these
scattered pieces to provide correct responses.

Our benchmark includes substantial coverage across this dimension: approximately 40\% of
test cases concentrate evidence in single messages for baseline evaluation, while 60\%
distribute information across multiple messages. Within the multi-message cases, we
maintain a gradient of complexity---from two-message scenarios that test basic connection
abilities to six-message challenges that require sophisticated information synthesis.
This distribution reveals critical performance boundaries, showing where systems
transition from reliable retrieval to struggling with scattered evidence.

Our validation mechanism ensures that questions can only be answered when all relevant
messages are present, preventing shortcuts and ensuring genuine multi-message reasoning.
This approach enables us to test not just whether a system can find information, but
whether it can assemble scattered evidence into coherent understanding.

With this foundational concept established, we now examine how each category tests
different memory capabilities, with both single and multi-message variants providing
complementary insights into system performance.

\subsubsection{User Facts}
User facts represents the foundational memory challenge---straightforward recall of
information the user has explicitly shared. This is the baseline capability upon
which all other memory functions build. User facts simply asks: can the system
remember what was told? This category forms the backbone of personalized assistance
by testing whether the AI can maintain basic continuity across conversations.

\textbf{Example (Single Message):}
\begin{itemize}
\item Evidence: ``My daughter Emma just got accepted to Stanford for computer science
  starting this fall.''
\item Question: ``What university is Emma attending?''
\item Answer: ``Stanford''
\end{itemize}

\textbf{Example (Multi-Message):}
\begin{itemize}
\item Evidence 1: ``I bought 50 shares of NVDA at \$420 each''
\item Evidence 2: ``I later bought another 30 shares of NVDA at \$380''
\item Evidence 3: ``Then I sold 20 shares when it hit \$500''
\item Question: ``How many NVDA shares do I currently own?''
\item Answer: ``60 shares''
\end{itemize}

Note: Multi-message examples for all other evidence categories are provided in Appendix C.

\subsubsection{Assistant Facts}
Assistant facts probes a different dimension: can the AI remember what it said?
While user facts tests whether the system recalls information provided by the user,
assistant facts evaluates whether the system can track and recall its own previous
statements and recommendations.

\textbf{Example:}
\begin{itemize}
\item Evidence (Assistant): ``For your team's remote collaboration, I recommend using
  Linear for issue tracking - it's much faster than Jira and has excellent keyboard
  shortcuts.''
\item Question: ``What project management tool did you recommend for my team?''
\item Answer: ``Linear''
\end{itemize}

\subsubsection{Abstention}
Abstention tests whether systems avoid hallucination when information is absent. This
category deliberately creates information gaps: conversations orbit around a topic
without providing the specific detail being asked about. The challenge isn't finding
the answer but recognizing its absence and admitting it rather than fabricating
plausible-sounding responses. In multi-message cases, multiple messages
discuss related topics without ever providing the specific answer, testing whether
the system can distinguish between having relevant context and having the actual answer.

\textbf{Example:}
\begin{itemize}
\item Evidence: ``John works in the marketing department and joined the company last March''
\item Question: ``What's John's phone number?''
\item Answer: ``There is no information in prior conversations to answer this question''
\end{itemize}

\subsubsection{Preferences}
Preferences move beyond facts to values---not just what the user has done but what they
like, how they work, what matters to them. This category tests whether the system can
build and apply a model of user preferences. The challenge isn't retrieving a specific
statement but understanding patterns: if someone values privacy and data control,
they probably won't appreciate a recommendation for cloud-based solutions.

\textbf{Example:}
\begin{itemize}
\item Evidence: ``I've been really enjoying React's component-based architecture and hooks
  API - it makes my frontend development so much cleaner''
\item Question: ``What framework would you recommend for building a dashboard?''
\item Answer (Rubric): ``A good response should recommend React-based tools and frameworks,
  emphasizing component reusability. Should NOT recommend other frameworks or vanilla
  approaches.''
\end{itemize}

\subsubsection{Changing Facts}
Changing facts tests how systems handle updates and corrections. This category evaluates
whether memory systems understand that new information invalidates the old rather than
simply accumulating all statements. In multi-message cases, changes accumulate across
messages, requiring the system to track all updates to determine the final state.

\textbf{Example:}
\begin{itemize}
\item Evidence 1: ``The client meeting is scheduled for Tuesday at 2 PM''
\item Evidence 2: ``Actually, we need to move the client meeting to Friday at 3 PM''
\item Question: ``When is the client meeting?''
\item Answer: ``Friday at 3 PM''
\end{itemize}

\subsubsection{Implicit Connections}
Implicit connections test the invisible threads of context---the unspoken considerations
that should shape every recommendation. A user mentions a broken ankle, then a week
later asks about weekend activities. They don't say ``remember my ankle''---that context
should already be active, coloring every suggestion. This category evaluates whether
memory systems can maintain ambient awareness of user circumstances and apply this
knowledge proactively without being reminded.

\textbf{Example:}
\begin{itemize}
\item Evidence: ``I slipped on ice last week and broke my ankle - I'm in a cast for the
  next 6 weeks''
\item Question (asked weeks later): ``What should I do this weekend to relax?''
\item Answer (Rubric): ``A good response should suggest sedentary activities. Should NOT
  suggest activities requiring mobility.''
\end{itemize}

Multi-message examples for all categories are provided in Appendix C.

\subsection{Benchmark Design Principles}

Our benchmark design prioritizes statistical rigor, model-agnostic evaluation, and
reproducibility through four core principles:

\subsubsection{Model Diversity for Universal Applicability}
We ensure model-agnostic evaluation by employing systematic diversity in both generation
and validation. Rather than relying on a single language model, we randomly select from
multiple model families (primarily Gemini and GPT-4) at each generation step, preventing
the benchmark from being tailored to any particular model's strengths. Validation requires
consensus across different architectures---each test case must be verified by multiple models
before acceptance. This approach ensures we test genuine memory capabilities rather than
model-specific patterns, providing fair evaluation across OpenAI, Google, Anthropic, and
open-source models.

\subsubsection{Scale and Statistical Power}
Our benchmark comprises 75,336 question-answer pairs, ensuring sufficient statistical
power for reliable evaluation. This represents a 150-fold increase over LongMemEval's
500 questions and provides category-level sample sizes that enable meaningful
statistical analysis.

\begin{table}[h]
\caption{Distribution of Test Cases Across Categories and Evidence Messages}
\centering
\begin{tabular}{lrrrrrrr}
\toprule
Category & 1 msg & 2 msg & 3 msg & 4 msg & 5 msg & 6 msg & Total \\
\midrule
User Facts & 7,642 & 4,405 & 2,170 & 1,063 & 802 & 651 & 16,733 \\
Assistant Facts & 10,000 & 911 & 886 & 797 & 135 & 16 & 12,745 \\
Changing Facts & - & 9,931 & 4,020 & 2,300 & 1,276 & 796 & 18,323 \\
Abstention & 10,000 & 2,952 & 1,958 & - & - & - & 14,910 \\
Preferences & 4,982 & 97 & - & - & - & - & 5,079 \\
Implicit Connections & 4,236 & 2,538 & 772 & - & - & - & 7,546 \\
\textbf{Total} & \textbf{36,860} & \textbf{20,834} & \textbf{9,806} & \textbf{4,160} & \textbf{2,213} & \textbf{1,463} & \textbf{75,336} \\
\bottomrule
\end{tabular}
\label{tab:test_cases_distribution}
\end{table}

This scale enables statistically significant results with confidence intervals under $\pm$2\%
for major categories while providing sufficient data for training memory-aware systems
through supervised learning.

\subsubsection{Coherent Synthetic Data Generation}
We ensure stylistic uniformity through a comprehensive pipeline that generates both filler
and evidence-containing conversations using identical methods. Unlike LongMemEval, which
sources filler conversations from external benchmarks creating detectable stylistic
differences, our unified approach prevents systems from achieving high scores by exploiting
stylistic patterns. The pipeline generates enterprise-focused personas spanning IT
administrators, financial analysts, and other professional roles, then creates conversations
through a three-phase process that balances diversity with quality control (detailed
implementation in Appendix A).

\subsubsection{Comprehensive Validation Framework}
Our multi-stage validation ensures that less than 5\% of generated examples pass all quality
checks, reflecting stringent standards while achieving massive scale. The fundamental
principle: we test memory systems, not reading comprehension. Questions must be answerable
with 100\% reliability when evidence is directly available, ensuring that performance
degradation in actual testing reflects memory retrieval challenges rather than ambiguous
questions or insufficient evidence. We employ multiple independent validation models and
require consecutive correct answers to eliminate lucky guesses. Different evidence categories
use tailored validation approaches---multi-message evidence requires all pieces to be essential,
abstention accepts ``I don't know'' as correct, and preference categories use rubric-based
evaluation (see Appendix A for complete validation protocols).

\subsubsection{Flexible Architecture and Extensibility}
The framework's modular design enables seamless integration and extension. Each evidence
category is self-contained---new memory challenges require only defining prompts and validation
rules without touching core infrastructure. This architecture proved its value when we
successfully converted both LongMemEval and LoCoMo to our format, demonstrating that existing
benchmarks can be unified under a common evaluation framework. The design supports variable
conversation lengths (2-300), different memory architectures (long context, RAG, hybrid),
and rapid prototyping of new categories (see Appendix A.3 for implementation details).

\subsection{Benchmark Execution Framework}

Our benchmark execution infrastructure addresses the practical challenges of running
large-scale memory evaluations, where a single comprehensive evaluation can cost hundreds
of dollars and take hours to complete. The framework provides a unified interface for all
memory systems---whether long context, RAG-based like Mem0, or hybrid block-based
approaches---enabling fair comparisons through standardized conversation loading and
question answering protocols. This abstraction layer ensures that each memory system
receives identical inputs and is evaluated using the same metrics, eliminating potential
biases from implementation differences.

The framework's sophisticated batch execution system divides test cases into balanced
batches ensuring equal representation across all context sizes (from 2 to 300 conversations),
preventing statistical skew that would occur if smaller contexts exhausted budgets before
larger contexts could be tested. Intelligent early termination reduces costs by 40-60\%
while maintaining statistical validity by monitoring convergence indicators. Multi-level
caching achieves 85-95\% hit rates for repeated conversation prefixes, dramatically reducing
evaluation costs. The framework's forensic capabilities proved invaluable---revealing, for
instance, that changing facts questions fail not because systems can't find updates but
because they struggle with temporal ordering when multiple changes occur. Complete
implementation details, including caching architecture, retry logic, and parallelization
strategies, are provided in Appendix A.4.

\section{From Zero to RAG: Progressive Memory Architectures for Growing Conversation Histories}
\label{sec:memory-rag}

This chapter establishes why conversational memory systems can achieve superior performance
through simple, ``naive'' approaches that would fail catastrophically in traditional RAG
systems. We begin by demonstrating that memory and RAG systems are fundamentally solving
the same problem---leveraging text corpora to enhance AI responses---and have evolved nearly
identical architectures to address shared challenges (Section \ref{subsec:convergence}). However, a critical
difference emerges: while RAG systems start with billions of tokens, memory systems begin
empty and grow progressively (Section \ref{subsec:progressive}). This unique characteristic enables memory
systems to employ computationally ``inefficient'' strategies like exhaustive search and full
context inclusion that deliver superior accuracy for the first 150+ conversations (Section
\ref{subsec:naive}). We then provide comprehensive experimental validation showing that naive long context
approaches achieve 70-82\% accuracy compared to 30-45\% for sophisticated RAG-based memory
systems (Section \ref{subsec:validation}), before discussing the implications for progressive architecture
strategies and the broader rethinking of memory research priorities (Section \ref{subsec:discussion}).

\subsection{The Memory-RAG Convergence}
\label{subsec:convergence}

Conversational memory and RAG share a fundamental objective: enhancing agent responses by
leveraging relevant context from a text corpus. Both domains face four critical challenges
that have driven remarkably parallel architectural evolution. \textbf{Temporal reasoning} requires
both to resolve ambiguous time references (``yesterday'' in conversations, ``recently'' in
documents) and manage information validity over time, leading both fields to develop explicit
timeline construction mechanisms \citep{arxiv:2505.01325, arxiv:2210.08750, arxiv:2505.20243}.
\textbf{Implicit information extraction} challenges both to resolve semantically incomplete inputs,
driving evolution from simple slot-filling to sophisticated reasoning frameworks that extract
and relate information fragments \citep{arxiv:2503.22458, arxiv:2509.10852, arxiv:2503.23095}.
\textbf{Knowledge update mechanisms} force both to handle contradictions and corrections, evolving
from recency heuristics to explicit versioning systems with REPLACE and DELETE operations
\citep{arxiv:2210.08750}. \textbf{Graph-structured representations} have emerged independently in both
domains---Zep and Mem0 for conversations, GraphRAG for documents---as the optimal architecture
for capturing entity relationships \citep{arxiv:2501.13956, arxiv:2504.19413, url:https://microsoft.github.io/graphrag/}.

This convergence is not coincidental but reflects fundamental requirements of context-aware
AI systems. The nearly identical solutions developed independently by both fields suggest
that advances in one domain should systematically transfer to the other. However, while the
problems and architectures converge, the characteristics of the underlying corpora diverge
dramatically---and this divergence, as we demonstrate below, enables memory systems to achieve
superior performance through approaches that would be impossible for traditional RAG (detailed
architectural analysis in Appendix B).

\subsection{Progressive Growth from Zero}
\label{subsec:progressive}

The most distinctive characteristic of conversational memory systems is their
progressive growth from an empty state. Unlike RAG systems that typically begin with
billions of tokens in their search space (enterprise documents, web content), memory
systems start with zero tokens and grow gradually through user interactions.

Consider a typical usage pattern: even if a user engages with an assistant for one hour
daily over four weeks, this generates only 100,000 tokens of conversation history---just 10\% of
the 1M token context windows available in current large language models. This fundamental
difference in scale creates opportunities for optimization strategies that would be
impractical in traditional RAG settings.

\subsection{Why Naive Policies Excel for Small Corpora}
\label{subsec:naive}

Operating on a much smaller search space enables the use of greedy and naive memory
retrieval policies optimized for this constrained environment. These strategies can
provide substantial benefits for memory systems while offering no advantage in RAG
scenarios with massive corpora. This asymmetry reveals unique research opportunities
specific to memory systems that deserve dedicated investigation.

Concretely, techniques that would be computationally prohibitive at web scale become
tractable for conversation-scale memory:
\begin{itemize}
\item Exhaustive search: With only hundreds rather than billions of text chunks, brute-force
  search becomes feasible
\item Complete reranking: Every retrieved item can be processed through expensive reranking
  models
\item Full attention mechanisms: Entire conversation histories can fit within transformer
  attention windows
\item Dynamic indexing: Indices can be rebuilt frequently as conversations evolve
\end{itemize}

This fundamental difference in scale---starting from zero and growing to perhaps thousands
of tokens rather than starting with billions---suggests that memory systems should embrace
rather than avoid ``inefficient'' approaches that prioritize accuracy over scalability.

In this paper, we demonstrate the surprising effectiveness of simply utilizing the entire
conversation history within long context windows. For a significant portion of the initial
user experience with an assistant, this brute-force approach can match or exceed the
performance of sophisticated retrieval mechanisms. Our empirical analysis reveals that long
context memory maintains 70-82\% accuracy on memory-dependent questions, compared to 30-45\%
for state-of-the-art RAG-based memory systems like Mem0, when operating on conversation
histories under 150 interactions. This substantial performance gap suggests that complexity
should be introduced progressively as conversation histories grow rather than deployed from
the outset.

\subsection{Experimental Validation}
\label{subsec:validation}

\subsubsection{Benchmark Performance Analysis}
We evaluate multiple memory architectures on our benchmark across varying conversation
history lengths, measuring accuracy, cost, and latency. Our analysis focuses on
conversations ranging from 0 to 300 prior interactions, representing typical usage
patterns before system transitions become necessary.

\subsubsection{Long Context Memory Performance}
Long context memory---placing all prior conversations directly in the model's
context---represents the simplest implementation yet consistently achieves the highest
accuracy. Our experiments reveal:

\textbf{Accuracy characteristics:} Long context maintains 82\% accuracy on memory-dependent
questions at the start, with gradual degradation to approximately 65\% at 300
conversations. This gentle decline contrasts sharply with more sophisticated systems
and aligns with findings from ``needle in the haystack'' literature on long context
performance \citep{arxiv:2507.22411}.

\begin{figure}[h]
\centering
\includegraphics[width=0.8\textwidth]{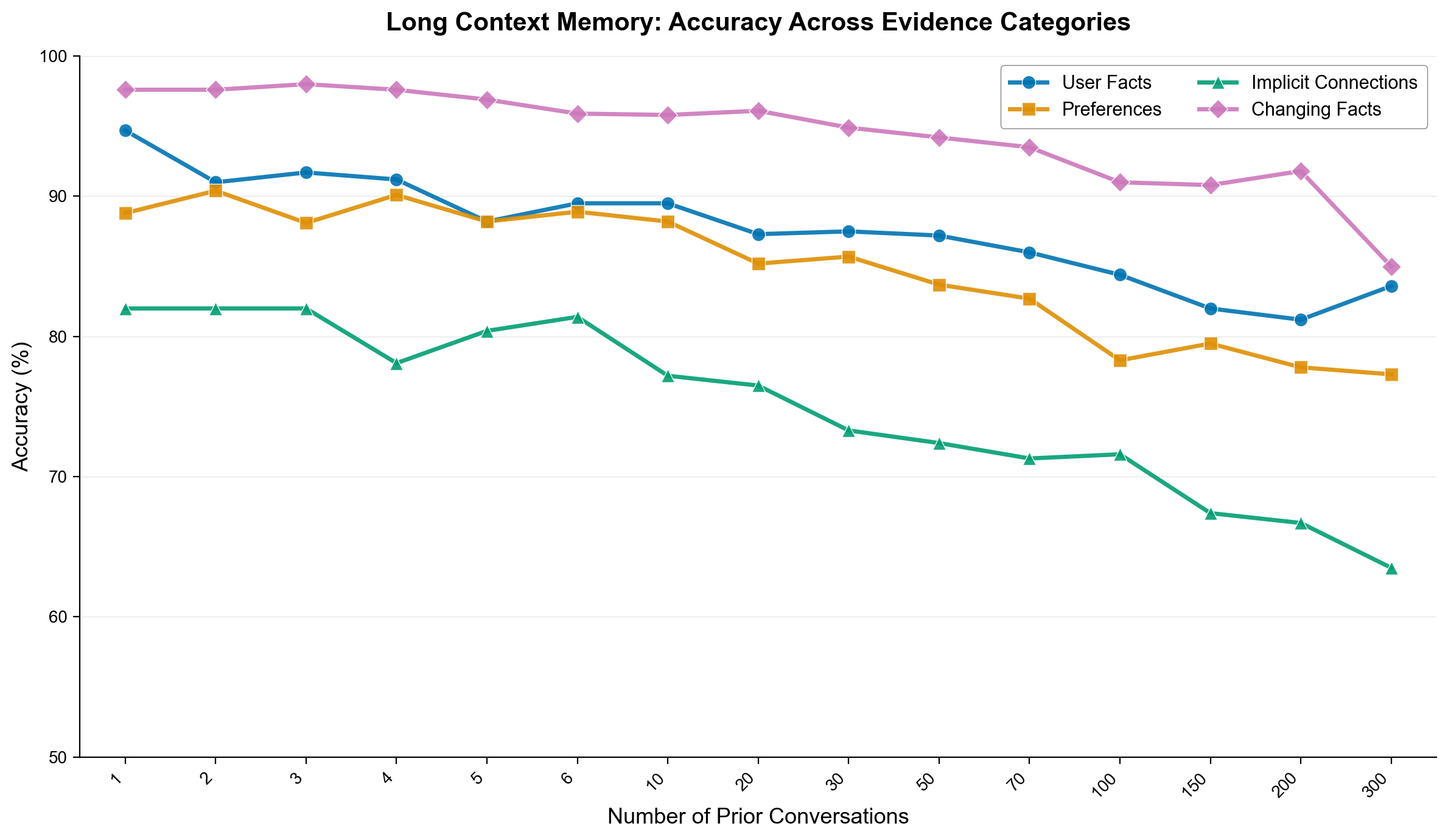}
\caption{Long Context Memory Accuracy}
\label{fig:long_context_accuracy}
\end{figure}

\begin{figure}[h]
\centering
\includegraphics[width=0.8\textwidth]{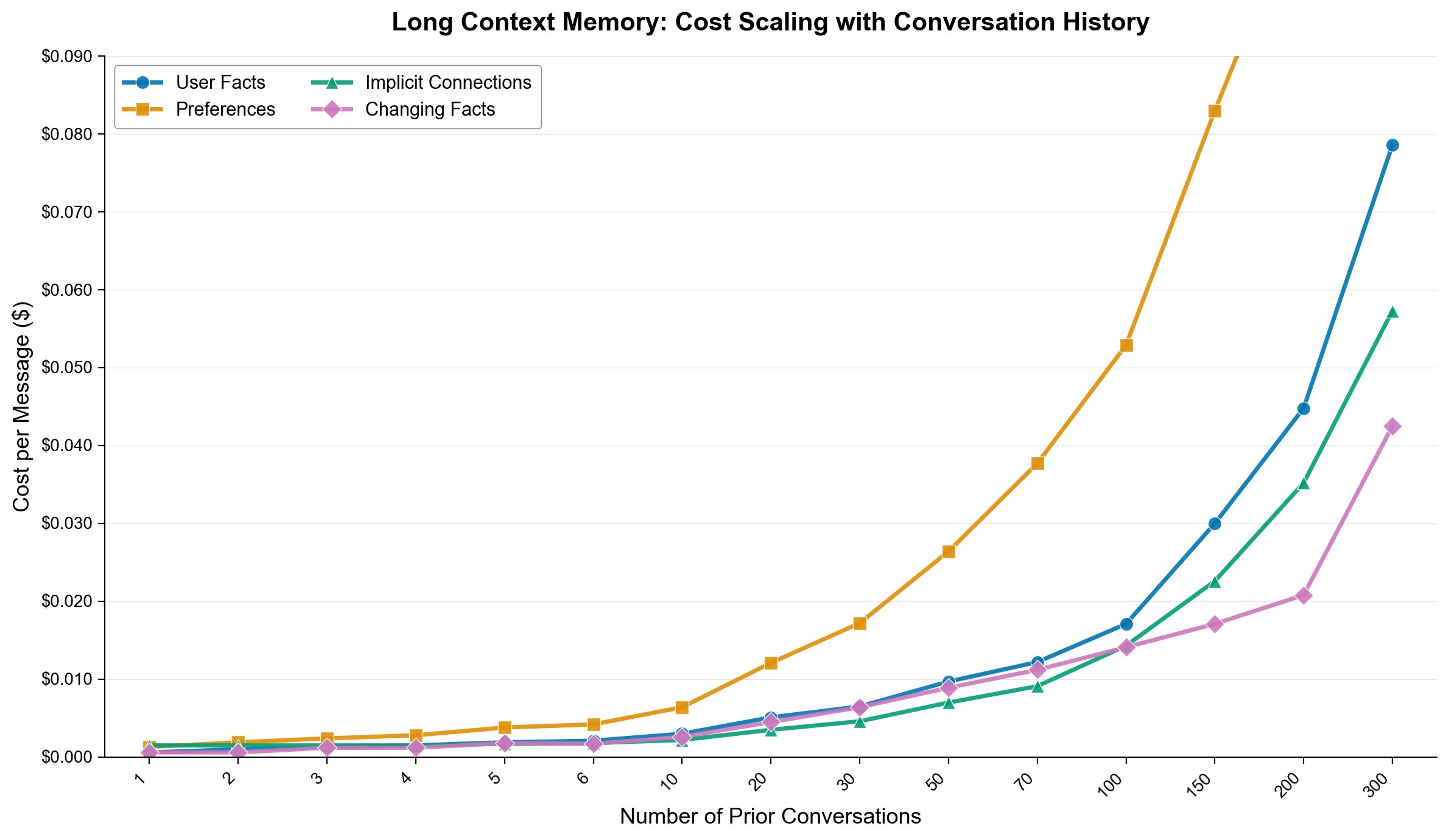}
\caption{Long Context Cost Scaling}
\label{fig:long_context_cost}
\end{figure}

\textbf{Cost scaling:} Message generation costs increase substantially with context size, as shown in Figure~\ref{fig:long_context_cost}.

Costs reach approximately \$0.08 per response at 300 conversations for User Facts in our
experimental setup (the cost to generate one AI response when answering a memory question).
While this represents significant cost growth, the actual impact depends heavily on the value
generated per interaction and will vary based on model choice and pricing.

\textbf{Latency patterns:} Response latency increases significantly with context size:

\begin{figure}[h]
\centering
\includegraphics[width=0.8\textwidth]{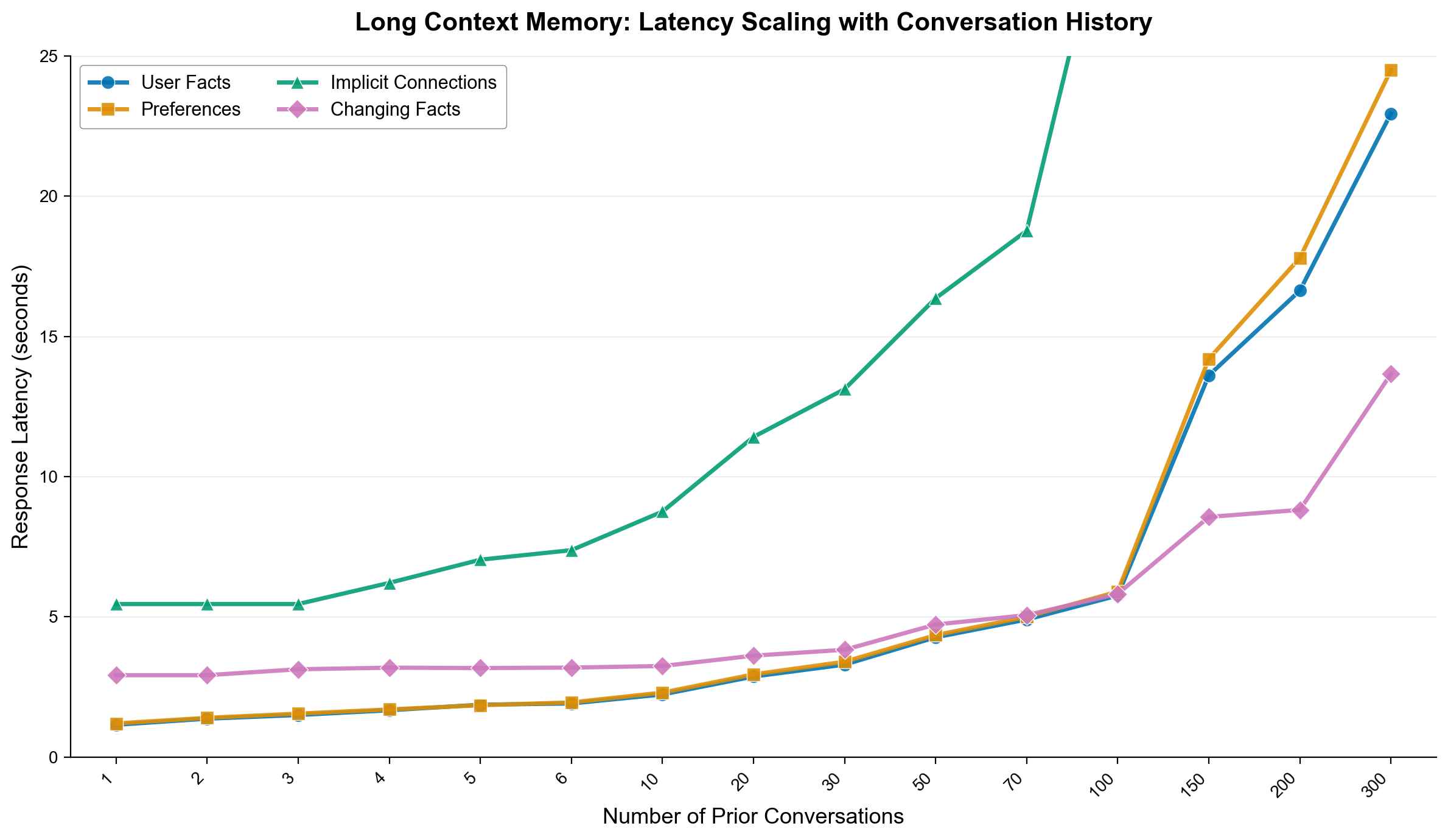}
\caption{Long Context Latency}
\label{fig:long_context_latency}
\end{figure}

\textbf{Test complexity scaling:} Examining how increasing the number of evidence items that must be
recalled affects accuracy across different context lengths:
\begin{figure}[h]
\centering
\includegraphics[width=0.8\textwidth]{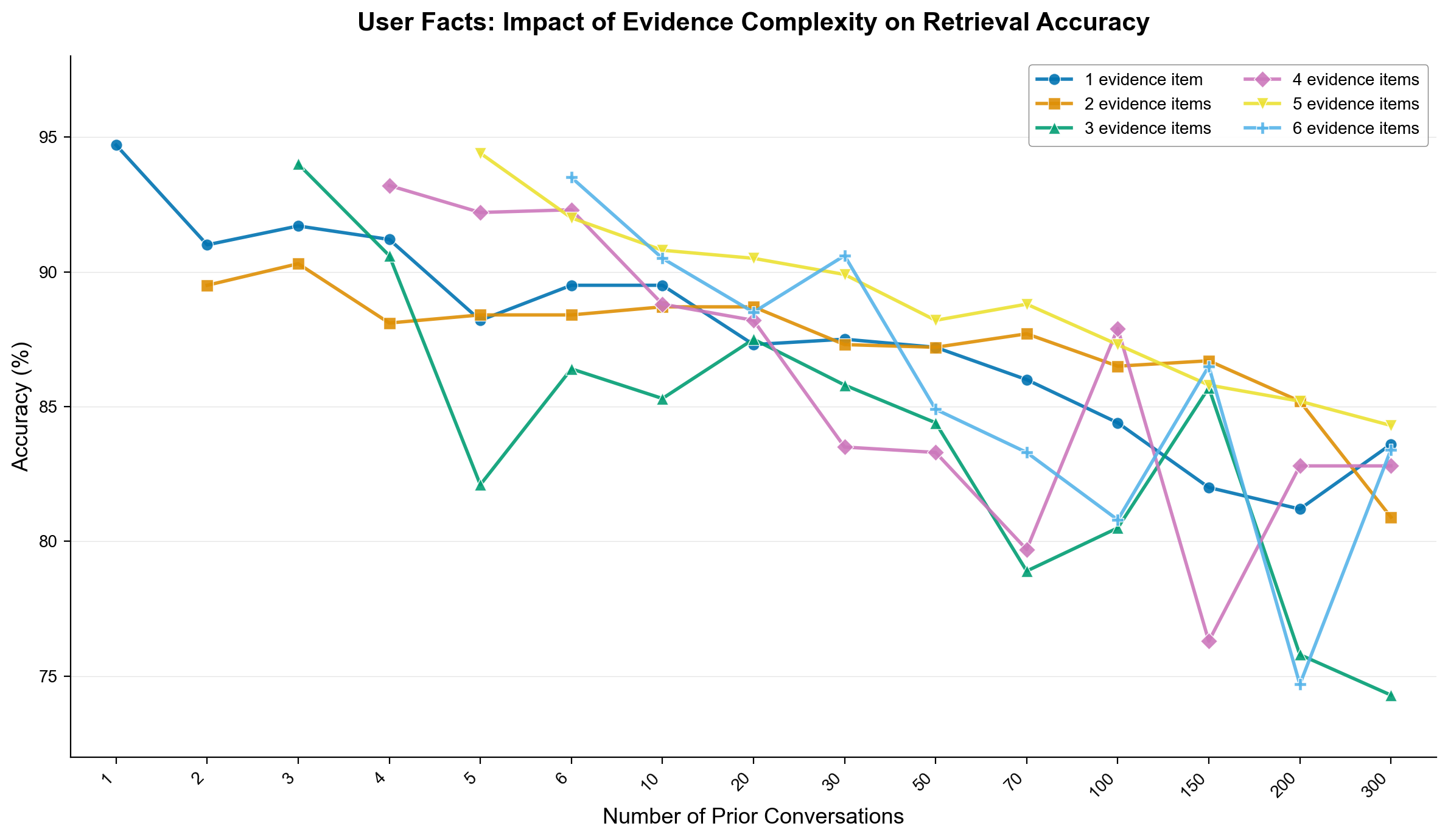}
\caption{User Facts Evidence Scaling}
\label{fig:user_facts_scaling}
\end{figure}

For User Facts, the number of evidence items (1-6) has virtually no impact on accuracy. All test
complexities show nearly identical degradation patterns from $\sim$95\% to 75-83\% at 300 conversations.

Changing Facts show relatively stable performance across evidence complexity levels, starting
at 96-98\% accuracy for small conversation histories but declining to 85-92\% at 300 conversations
depending on the number of evidence items tracked.

\begin{figure}[h]
\centering
\includegraphics[width=0.8\textwidth]{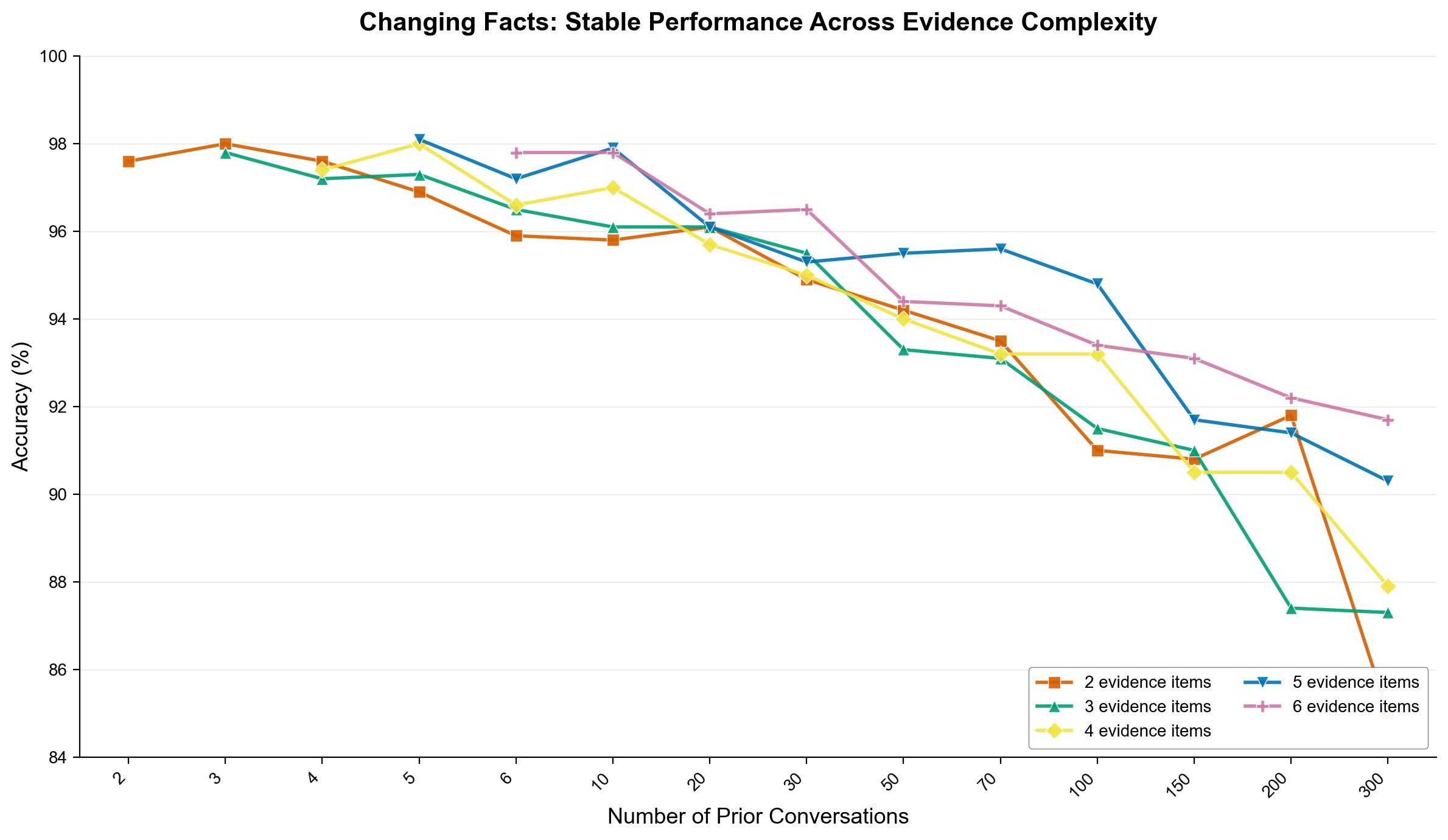}
\caption{Changing Facts Evidence Scaling}
\label{fig:changing_facts_scaling}
\end{figure}

The explicit nature of change statements
(``I now prefer X instead of Y'') helps maintain higher accuracy than static facts, though
performance still degrades as conversation history grows.

\textbf{Key insight from all performance analyses:} Combining the evidence from accuracy, cost,
latency, and evidence scaling charts above, we find that for conversation histories under
30-50 interactions, long context provides the best accuracy with acceptable cost and latency
characteristics, making it the optimal choice for initial user interactions.

\subsubsection{Model Size Analysis for Long Context Memory}
Our comprehensive evaluation of long context memory performance across three model tiers---Gemini 2.5
Flash Lite, Flash, and Pro---reveals a non-linear relationship between model size and memory
performance, identifying a clear optimal range for long context-based conversational memory systems.

\textbf{Comparing Flash vs Pro: Diminishing Returns at Scale}
\begin{figure}[h]
\centering
\includegraphics[width=0.8\textwidth]{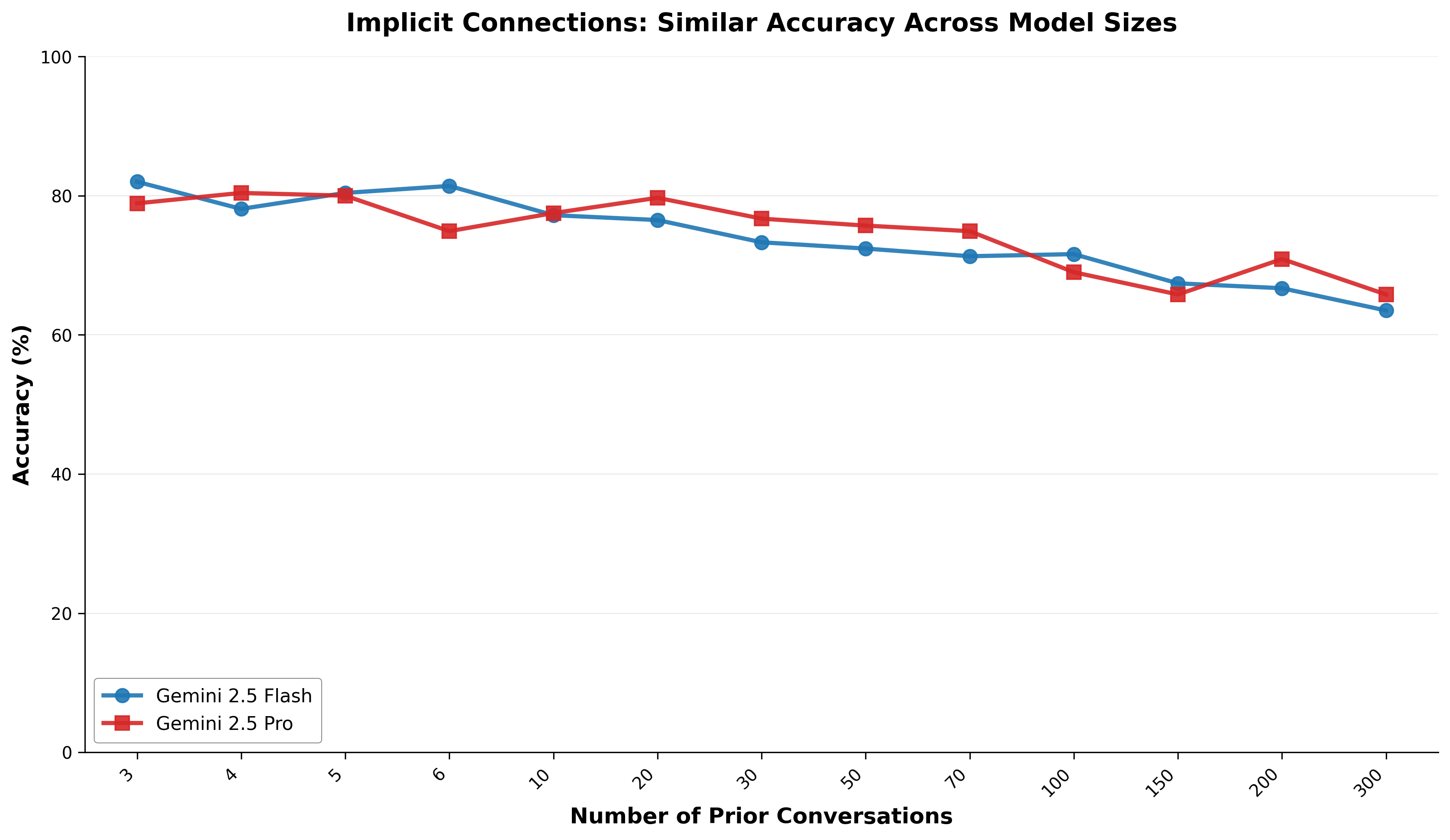}
\caption{Implicit Connections: Flash vs Pro}
\label{fig:implicit_flash_vs_pro}
\end{figure}

For implicit connections requiring contextual reasoning, Flash and Pro demonstrate nearly identical
performance trajectories. At 300 conversations, the accuracy difference is only 2.3 percentage points
(63.5\% Flash vs 65.8\% Pro), with performance curves overlapping throughout most of the conversation
range. This minimal gap suggests that memory recall capabilities plateau at mid-tier model sizes.

Even in challenging multi-fact scenarios with 6 evidence items, Pro's advantage remains marginal.
The 6 percentage point difference at 300 conversations (83.4\% Flash vs 89.4\% Pro) represents less
than 10\% relative improvement, indicating that additional model capacity provides limited benefit
for memory-specific tasks.

\begin{figure}[h]
\centering
\includegraphics[width=0.8\textwidth]{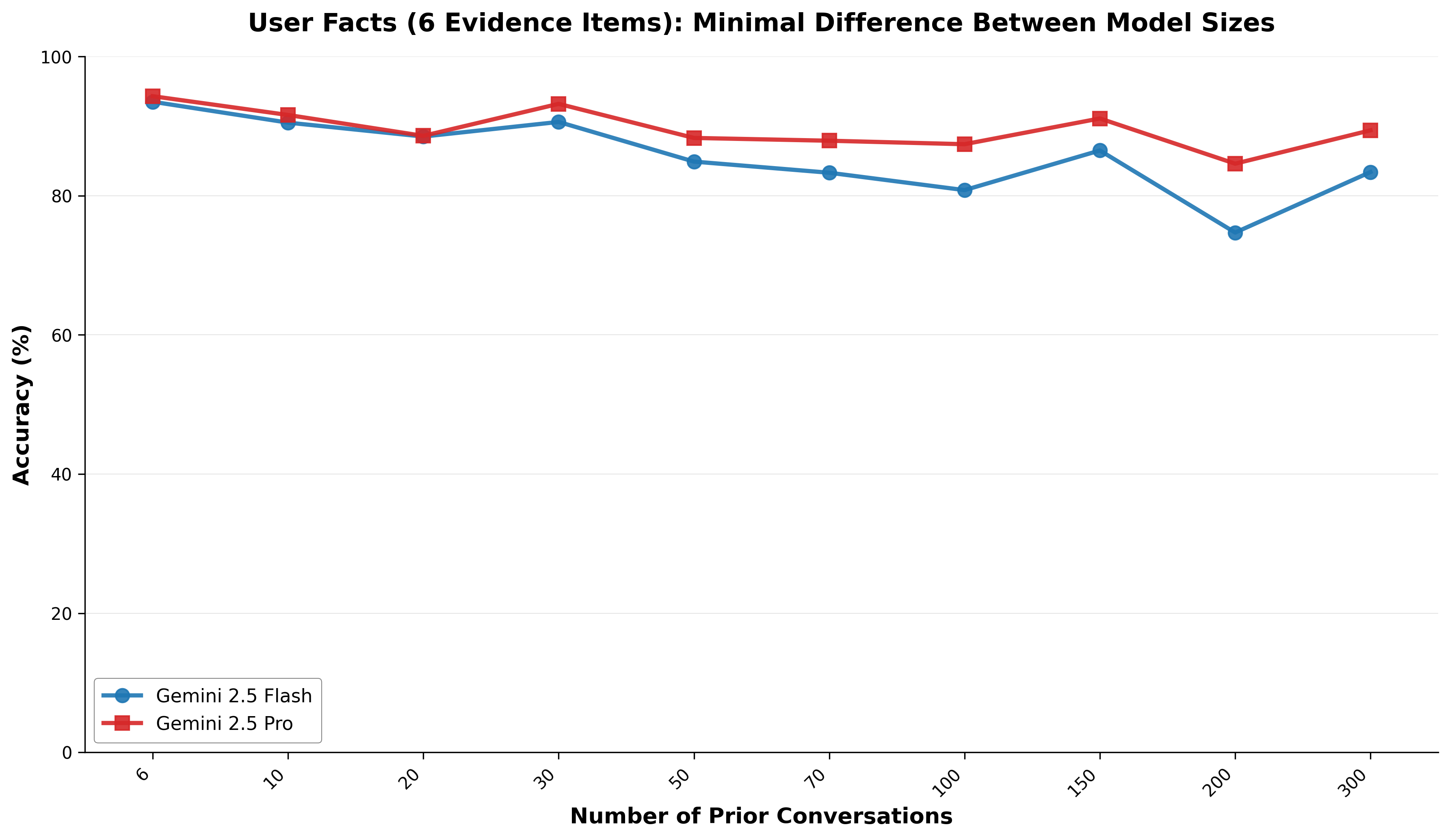}
\caption{User Facts: Flash vs Pro}
\label{fig:user_facts_flash_vs_pro}
\end{figure}

\begin{figure}[h]
\centering
\includegraphics[width=0.8\textwidth]{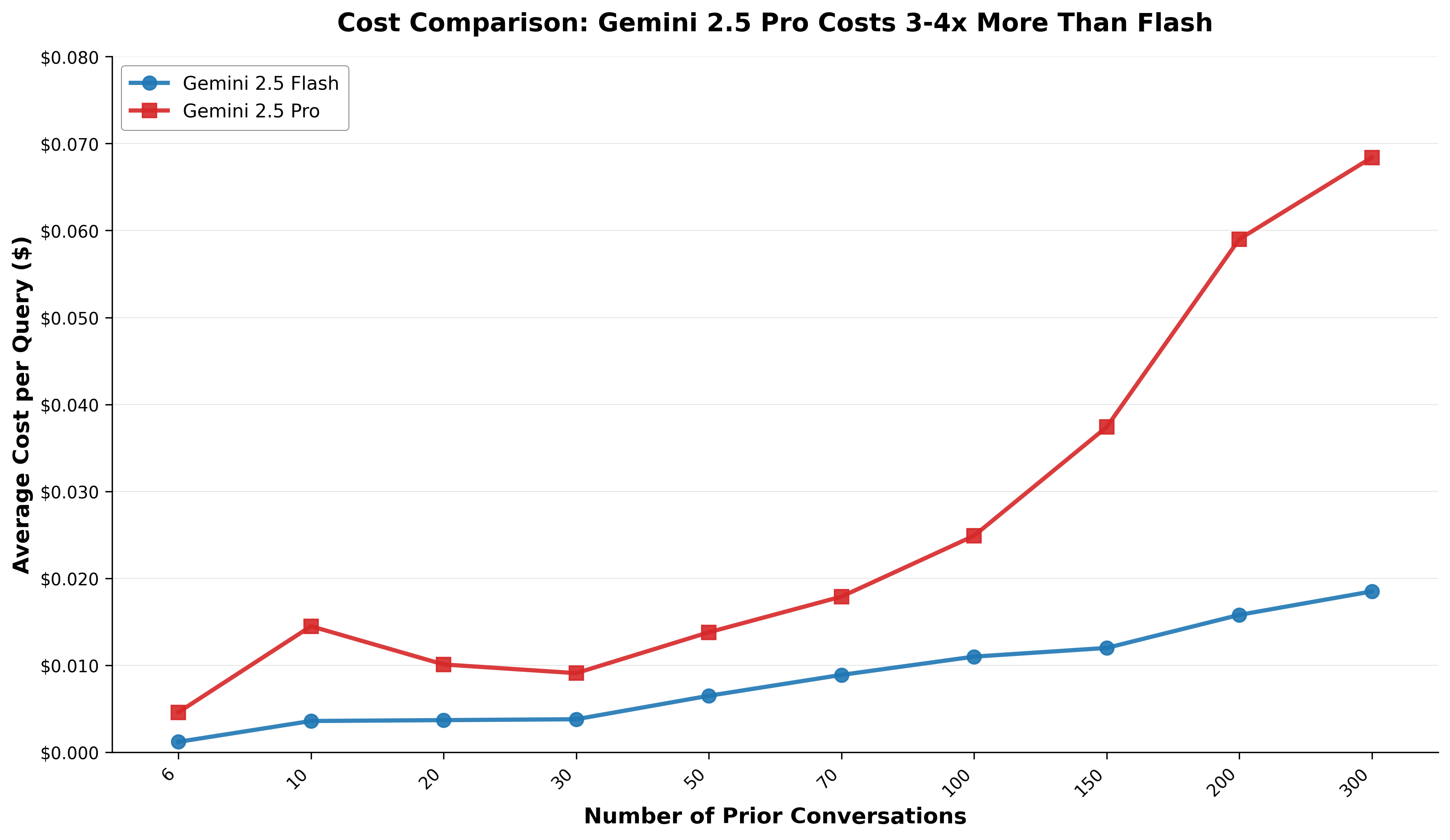}
\caption{Cost Analysis: Flash vs Pro}
\label{fig:cost_flash_vs_pro}
\end{figure}

The economic analysis reveals a stark trade-off: Pro costs 3.7x more than Flash at 300 conversations
(\$0.0684 vs \$0.0185), with the cost ratio increasing as conversation history grows. This substantial
cost premium for marginal accuracy gains makes Pro economically inefficient for most deployments.

\textbf{Comparing Flash vs Flash Lite: Performance Floor for Memory Tasks}
\begin{figure}[h]
\centering
\includegraphics[width=0.8\textwidth]{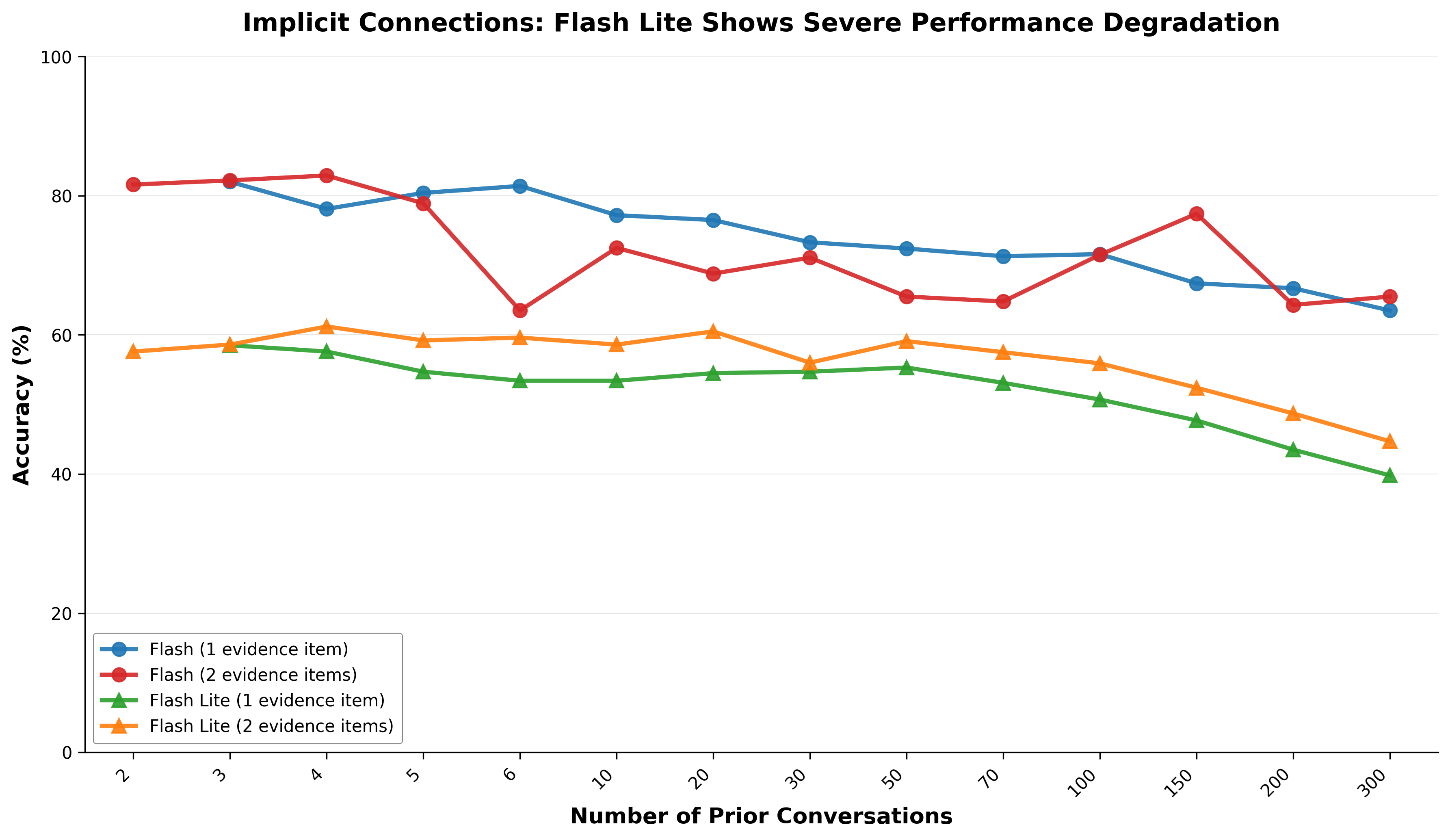}
\caption{Implicit Connections: Flash vs Flash Lite}
\label{fig:implicit_flash_vs_lite}
\end{figure}

In contrast to the Flash-Pro comparison, reducing model size to Flash Lite results in significant
performance degradation. For implicit connections with single evidence items, accuracy drops from
63.5\% to 39.8\% at 300 conversations---a 24 percentage point reduction. With 2 evidence items, both
models show similar degradation patterns, but Flash Lite consistently underperforms by 15-20
percentage points across all conversation sizes.

Multi-fact scenarios reveal further limitations of ultra-lightweight models. With 6 evidence items,
Flash achieves 83.4\% accuracy at 300 conversations while Flash Lite reaches only 52.2\%---a 31
percentage point difference. This performance level indicates that nearly half of facts are
not correctly recalled in complex scenarios.

\begin{figure}[h]
\centering
\includegraphics[width=0.8\textwidth]{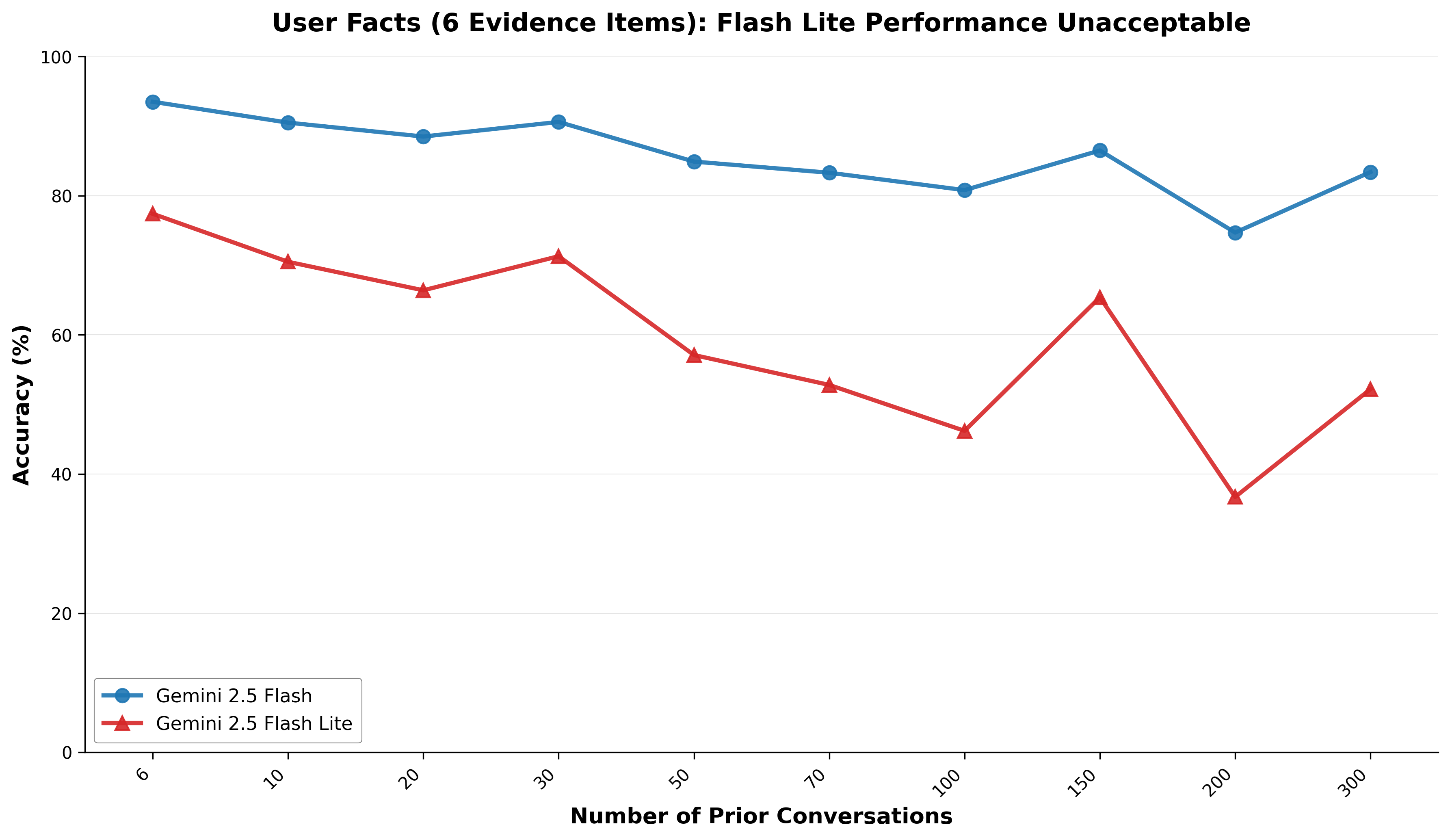}
\caption{User Facts: Flash vs Flash Lite}
\label{fig:user_facts_flash_vs_lite}
\end{figure}

\textbf{Key findings:} Our analysis identifies mid-tier models (Flash-class) as the optimal choice for
long context memory systems. When using long context approaches, these models operate at a sweet spot where:
\begin{enumerate}
\item Memory performance has reached effective saturation (within 2-6\% of larger models)
\item Costs remain manageable (less than 30\% of premium model costs)
\item Accuracy stays above critical thresholds (>80\% for most evidence types)
\end{enumerate}

For long context memory systems, moving up to larger models yields diminishing returns---minimal
accuracy improvements at disproportionate cost increases. Moving down to ultra-lightweight models
crosses a performance threshold where accuracy degrades substantially, particularly for complex
memory tasks. This non-linear scaling pattern suggests that long context memory requires a minimum
model capacity to effectively process and recall information from extended conversation histories,
above which additional parameters provide limited benefit.

\subsubsection{RAG-Based Memory System Performance}
We evaluated Mem0 as a representative example of sophisticated RAG-style memory systems
that index conversations and create structured representations for retrieval. The results
reveal significant performance trade-offs that vary dramatically by evidence type:

\textbf{User Facts Performance:}
\begin{figure}[h]
\centering
\includegraphics[width=0.8\textwidth]{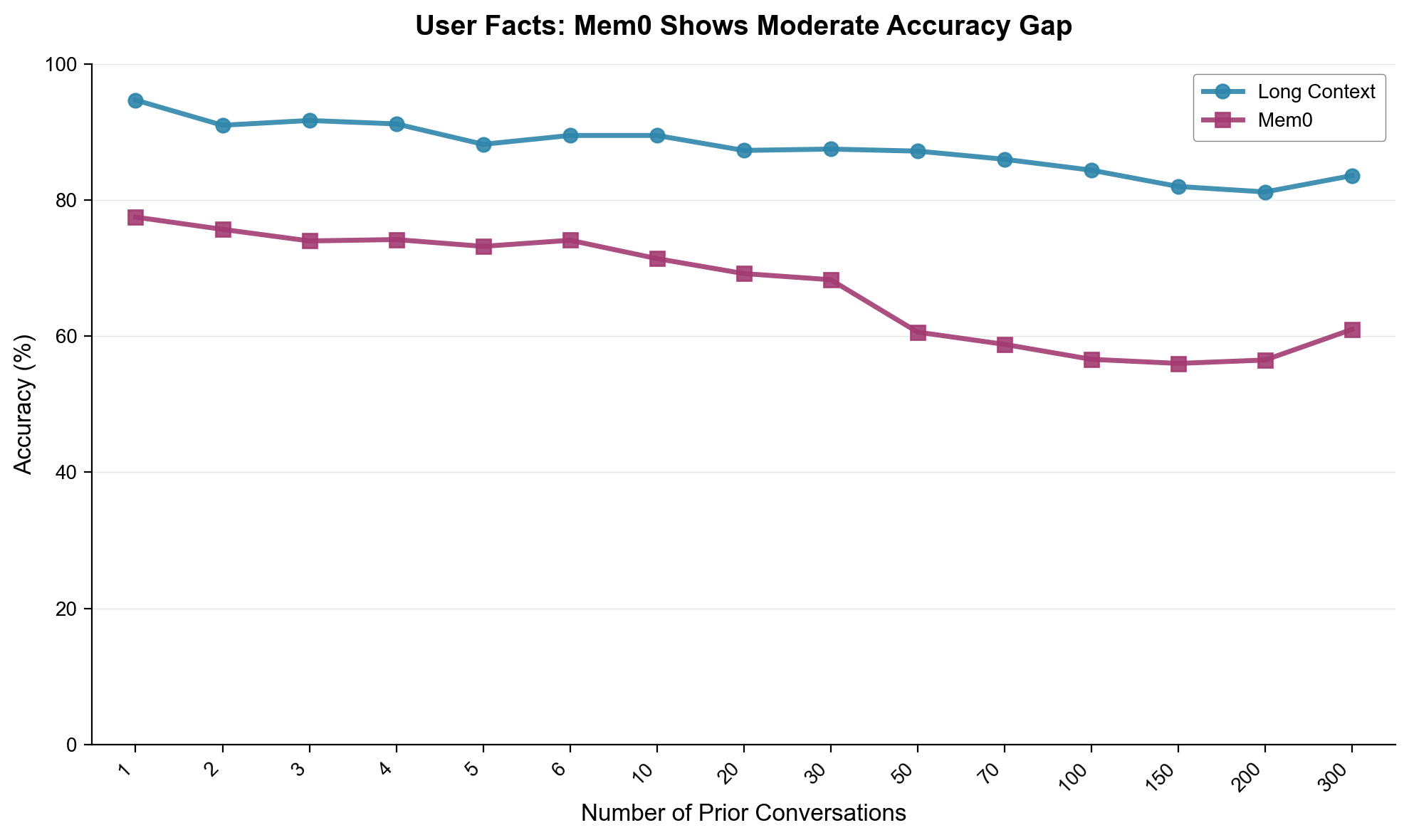}
\caption{User Facts: Mem0 vs Long Context}
\label{fig:mem0_user_facts}
\end{figure}

For straightforward factual recall, Mem0 shows a moderate but consistent accuracy gap
compared to long context. Starting at 77.5\% accuracy versus 94.7\% for long context, the
gap remains relatively stable at 15-25 percentage points across all conversation scales.
This indicates that while Mem0 can retrieve basic facts, it misses approximately one in
four facts that long context successfully recalls.

\textbf{Preferences \& Implicit Connections Performance:}

\begin{figure}[h]
\centering
\includegraphics[width=0.8\textwidth]{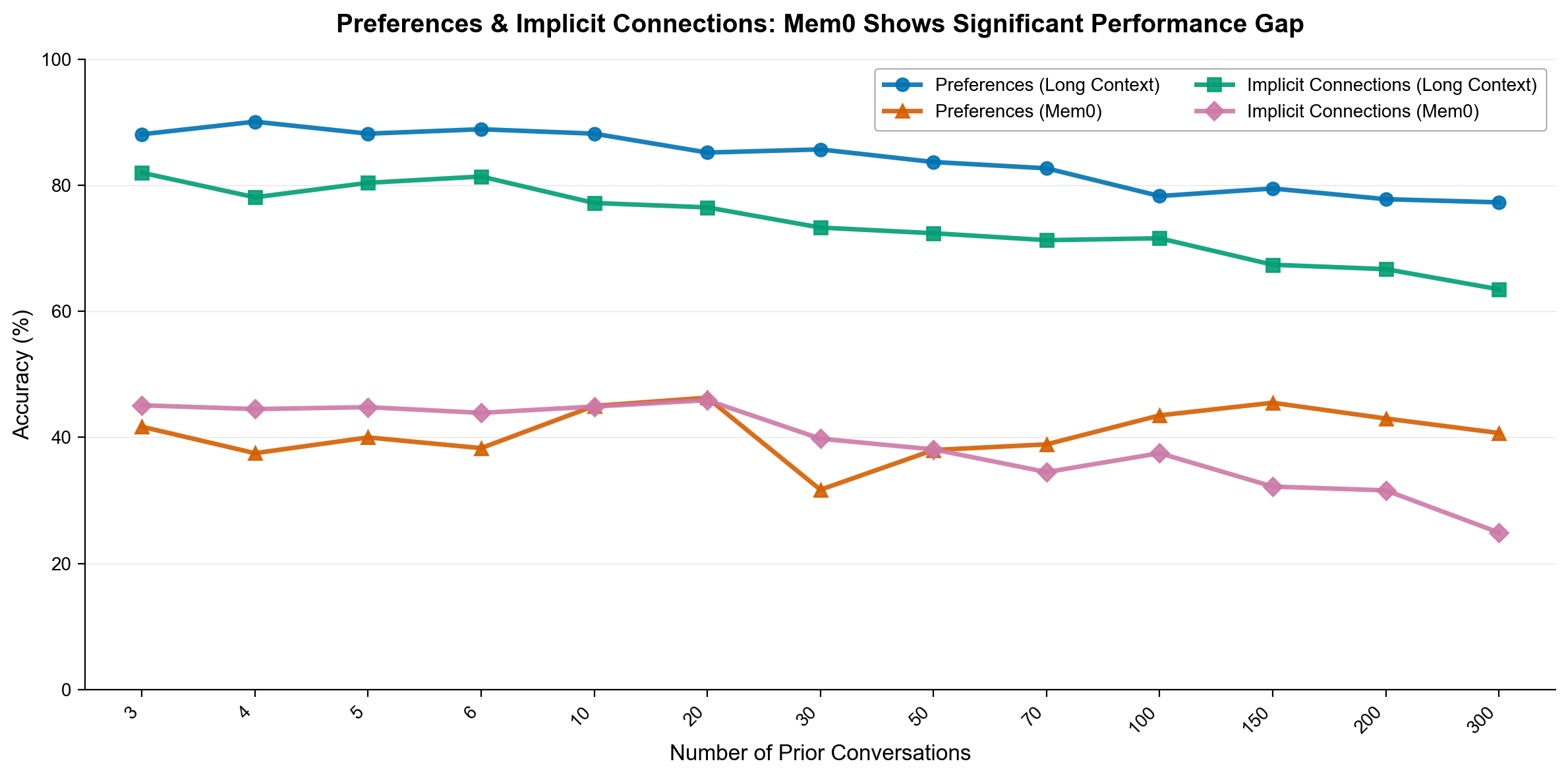}
\caption{Preferences \& Implicit: Mem0 vs Long Context}
\label{fig:mem0_preferences_implicit}
\end{figure}

The performance gap becomes dramatic for evidence types requiring nuanced understanding.
For preferences, Mem0 achieves only 30-45\% accuracy compared to 77-90\% for long context,
indicating that the system fails to retrieve more than half of user preferences.
Implicit connections fare similarly poorly, with Mem0 achieving 25-45\% accuracy versus
63-82\% for long context. These results suggest that current RAG architectures fundamentally
struggle with contextual reasoning and implicit information that isn't explicitly stated.

\textbf{Cost and Latency Comparison:}

\begin{figure}[h]
\centering
\includegraphics[width=0.8\textwidth]{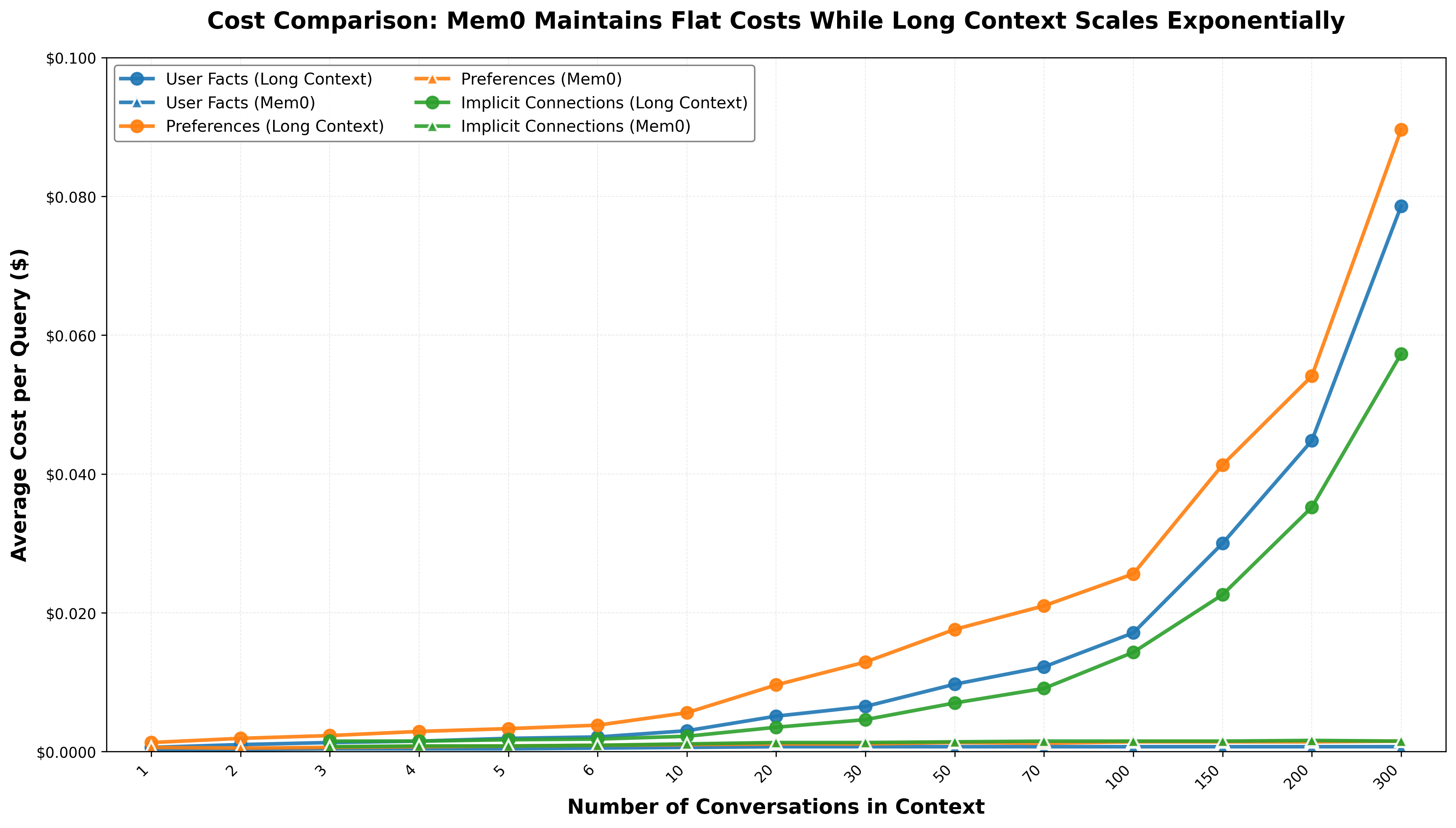}
\caption{Cost Comparison: Mem0 vs Long Context}
\label{fig:mem0_cost}
\end{figure}

The cost advantage of Mem0 becomes overwhelming at scale. While Long Context costs grow
dramatically from \$0.001 to \$0.09 per query at 300 conversations, Mem0 maintains relatively
stable costs around \$0.0007-\$0.0015 with minimal variation across history lengths---achieving
up to 95x cost reduction at scale. This dramatic difference makes Mem0 economically viable
for production deployments where Long Context would be prohibitively expensive.

Response time analysis for User Facts reveals interesting dynamics.

\begin{figure}[h]
\centering
\includegraphics[width=0.8\textwidth]{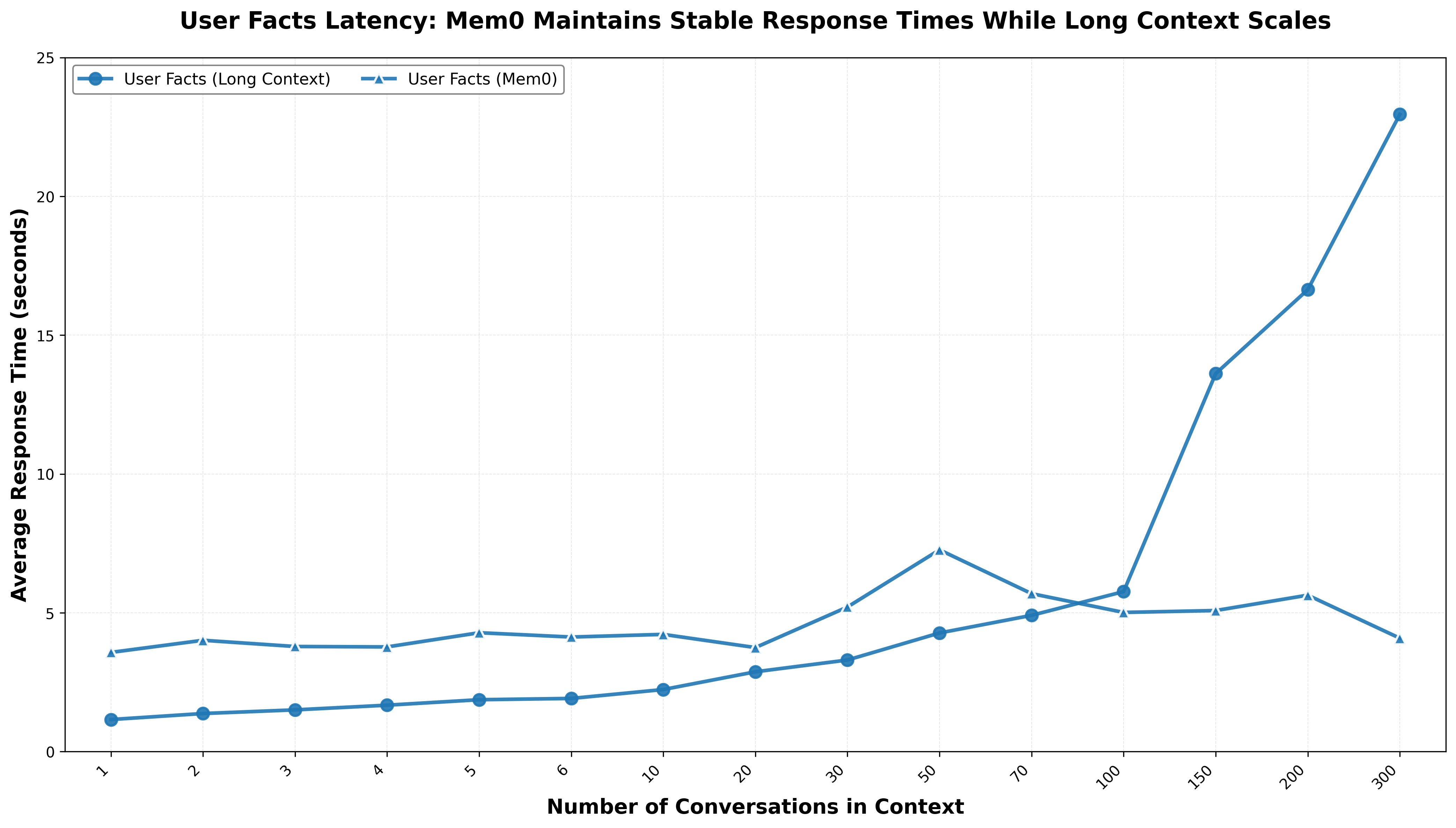}
\caption{User Facts Latency: Mem0 vs Long Context}
\label{fig:mem0_latency}
\end{figure}

Initially, Mem0 shows
higher baseline latency (3-4 seconds) compared to Long Context (1-2 seconds) due to the
overhead of retrieval operations. However, as conversation history grows, Long Context
latency increases steadily to 23 seconds at 300 conversations, while Mem0 maintains stable
3-7 second response times. The crossover point occurs around 10-20 conversations, after which
Mem0 becomes consistently faster despite its accuracy limitations.

\textbf{Evidence Scaling Impact on Mem0:}
\begin{figure}[h]
\centering
\includegraphics[width=0.8\textwidth]{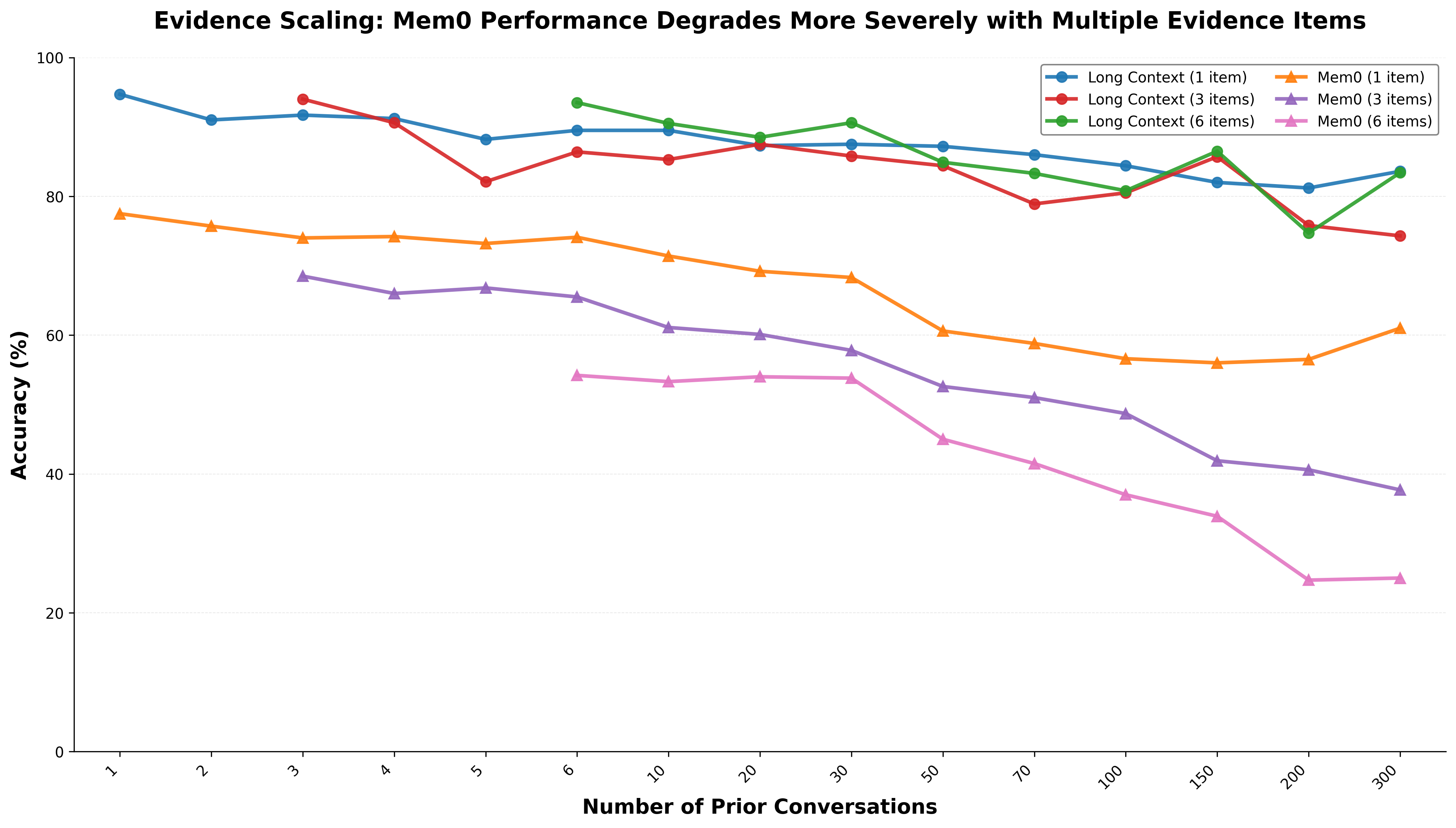}
\caption{Evidence Scaling: Mem0 vs Long Context}
\label{fig:mem0_evidence_scaling}
\end{figure}

A critical limitation of Mem0 emerges when handling multiple evidence items. While Long Context
maintains relatively stable accuracy across 1, 3, and 6 evidence items (declining from $\sim$84\% to
$\sim$83\% at 300 conversations), Mem0 shows severe degradation. With a single evidence item,
Mem0 achieves 61\% accuracy at 300 conversations---a 23-point gap. However, with 3 items,
accuracy drops to 38\% (37-point gap), and with 6 items, it decreases to 25\% (58-point gap).
This suggests that RAG architectures struggle fundamentally with multi-fact retrieval scenarios,
where relevant information is distributed across multiple memory entries.

\textbf{Key insight from Mem0 performance analysis:} Combining the evidence from accuracy, cost,
latency, and evidence scaling comparisons above, we find that Mem0 becomes the default choice
for conversation histories exceeding 50-100 interactions despite its accuracy limitations.
The 95x cost reduction and predictable sub-10 second latencies make Long Context economically
infeasible at scale. However, Mem0's severe limitations---15-25 point accuracy gaps for single
facts, 75\% accuracy reduction with 6 evidence items, and poor performance on nuanced tasks---mean
that applications must be designed around these constraints. This creates
a fundamental trade-off: beyond 100 conversations, you must either accept degraded memory
performance with Mem0 or face prohibitive costs with Long Context.

\subsubsection{Hybrid Approaches: Two-Phase Extraction Architecture}
To address the scalability limitations of long context while maintaining higher accuracy
than pure RAG systems, we developed hybrid extraction approaches that combine the benefits
of both paradigms. These approaches employ a two-phase architecture that separates information
extraction from answer generation, enabling different model choices for each phase:

\textbf{Phase 1 - Extraction:} A model processes conversations to extract relevant information

\textbf{Phase 2 - Answering:} A model generates the final answer using only the extracted information

We implemented and evaluated two distinct extraction strategies:

\paragraph{Block-Based Extraction Strategy}

Block-based extraction divides conversations into blocks of 10, extracts relevant information
from each block independently, then aggregates the extracted information for final answer generation.

\begin{figure}[h]
\centering
\includegraphics[width=0.8\textwidth]{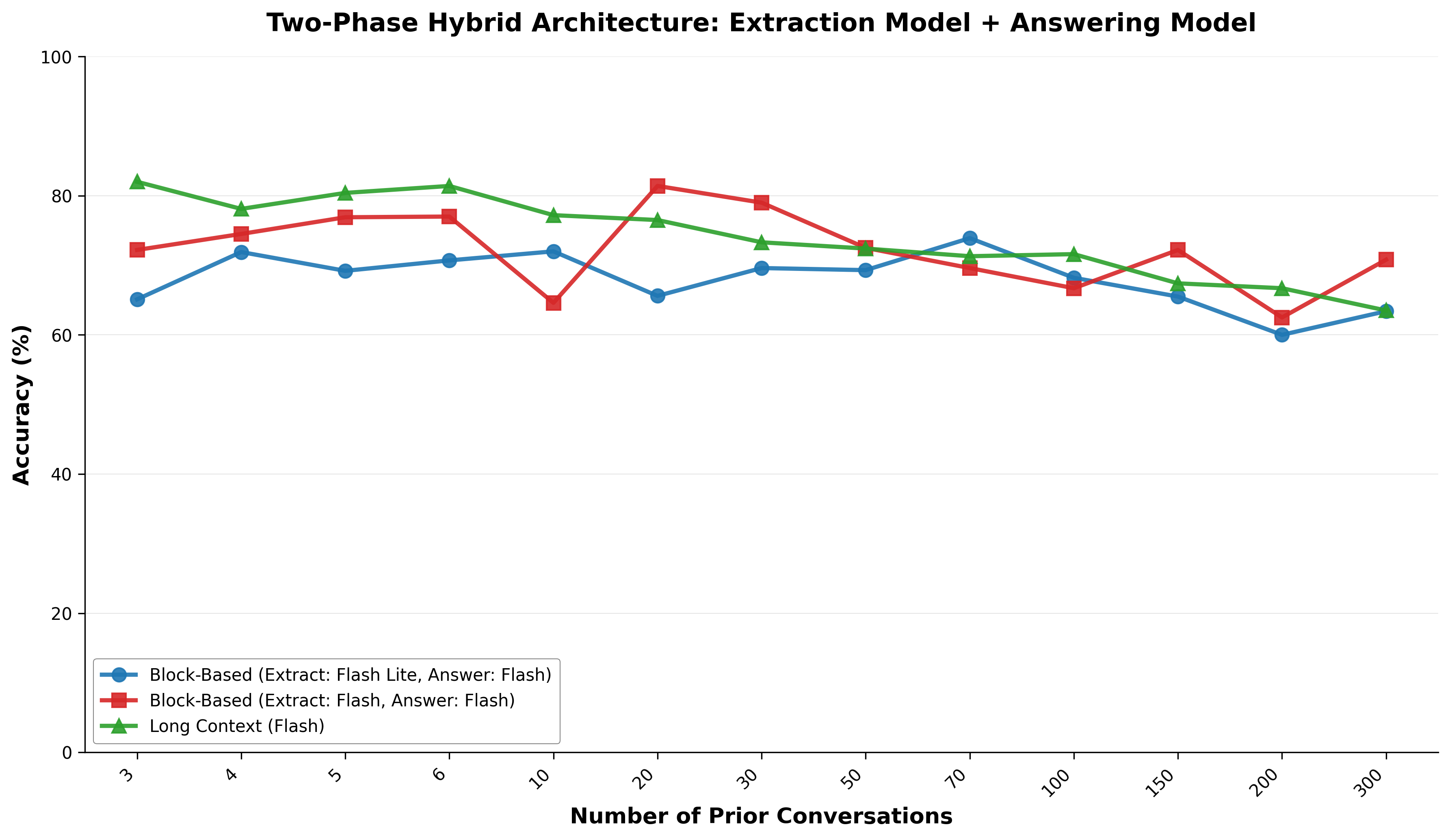}
\caption{Block-Based Extraction Performance (Implicit Connection, 1 Evidence Item)}
\label{fig:block_lightweight}
\end{figure}

The chart demonstrates three configurations for block-based extraction on implicit connection tasks
with 1 evidence item:

\begin{enumerate}
\item \textbf{Block-Based (Extract: Flash, Answer: Flash)} - Red line: Using Flash for both phases
   achieves 70.8\% accuracy at 300 conversations, outperforming pure long context (63.5\%) by
   7.3 percentage points. This improvement suggests that structured extraction reduces context
   confusion compared to processing all conversations directly.

\item \textbf{Block-Based (Extract: Flash Lite, Answer: Flash)} - Blue line: Using the lightweight
   Flash Lite for extraction while maintaining Flash for answering achieves 63.4\% accuracy,
   nearly matching the long context baseline. This demonstrates that for implicit connection
   tasks, a smaller extraction model can maintain performance while reducing computational costs.

\item \textbf{Long Context (Flash)} - Green line: Traditional approach using Flash to process the
   entire conversation history directly, serving as our baseline at 63.5\% accuracy.
\end{enumerate}

The block-based approach reveals a nuanced cost optimization opportunity: using Flash Lite for
extraction maintains comparable accuracy (63.4\% vs 63.5\% baseline) while reducing per-token costs
by approximately 10x compared to Flash. However, this approach requires 30 separate API calls (one
per block) versus a single call for long context, introducing fixed request overhead that can offset
token savings at smaller scales. For implicit connection tasks at 300 conversations, the substantial
token volume makes Flash Lite extraction economically favorable despite request overhead. At shorter
conversation histories (under 50 conversations), the multiple API calls may actually increase total
costs compared to single-request long context, requiring careful consideration of the break-even point
for each deployment scenario.

\textbf{Performance with Increased Evidence Complexity:}
\begin{figure}[h]
\centering
\includegraphics[width=0.8\textwidth]{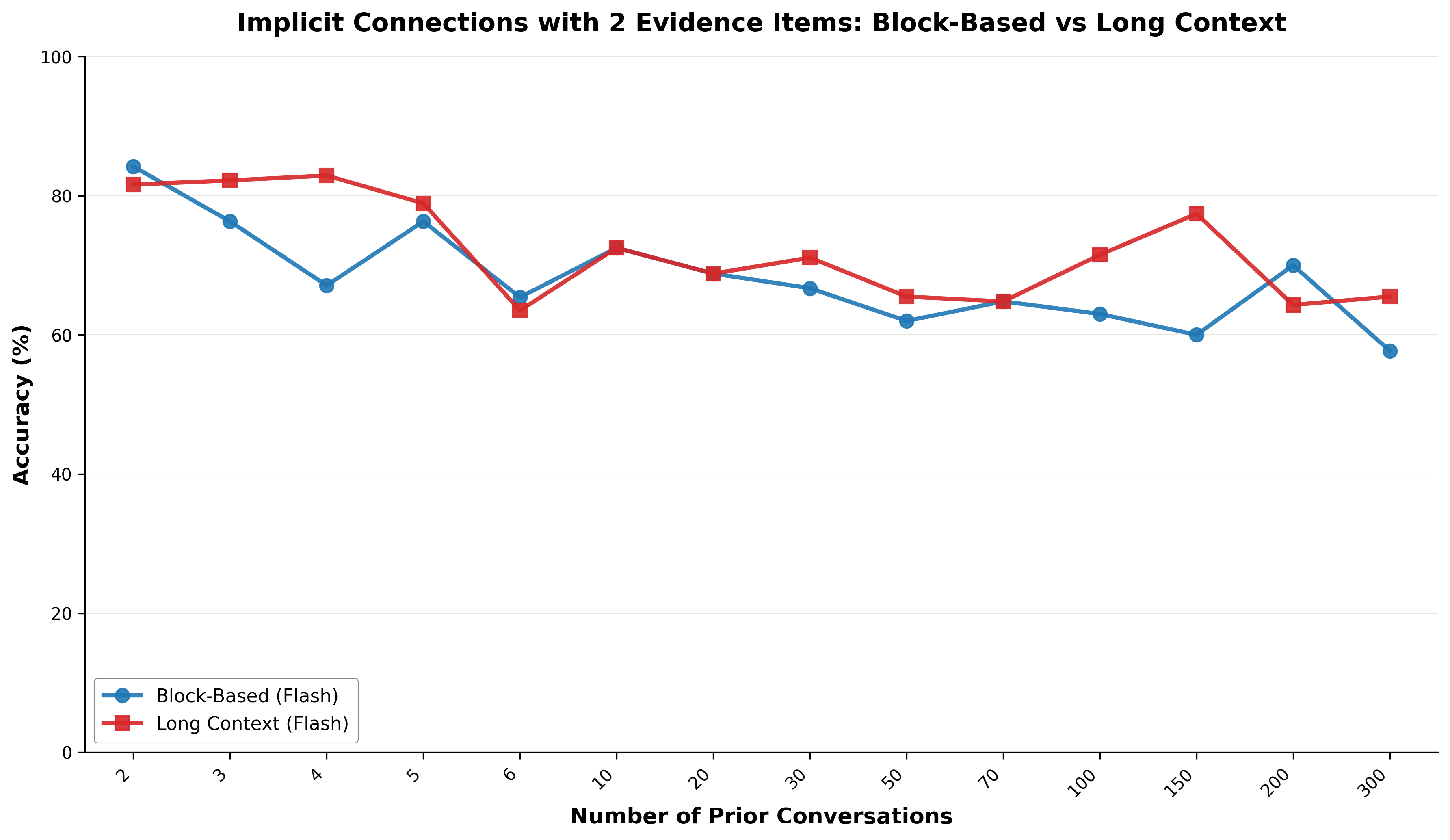}
\caption{Block-Based vs Long Context (Implicit Connection, 2 Evidence Items)}
\label{fig:block_two_evidence}
\end{figure}

When task complexity increases to 2 evidence items, the performance dynamics shift. At 300 conversations,
long context (65.5\%) outperforms block-based extraction (57.7\%) by 7.8 percentage points. However,
at smaller conversation scales (2-10 conversations), block-based extraction maintains parity or slight
advantages. This pattern suggests that block-based extraction excels at focused extraction from smaller
contexts but faces challenges aggregating multiple pieces of evidence across many blocks. The crossover
point around 30-50 conversations indicates where the benefits of structured extraction are offset by
the complexity of coordinating multiple evidence items across blocks.

\begin{figure}[h]
\centering
\includegraphics[width=0.8\textwidth]{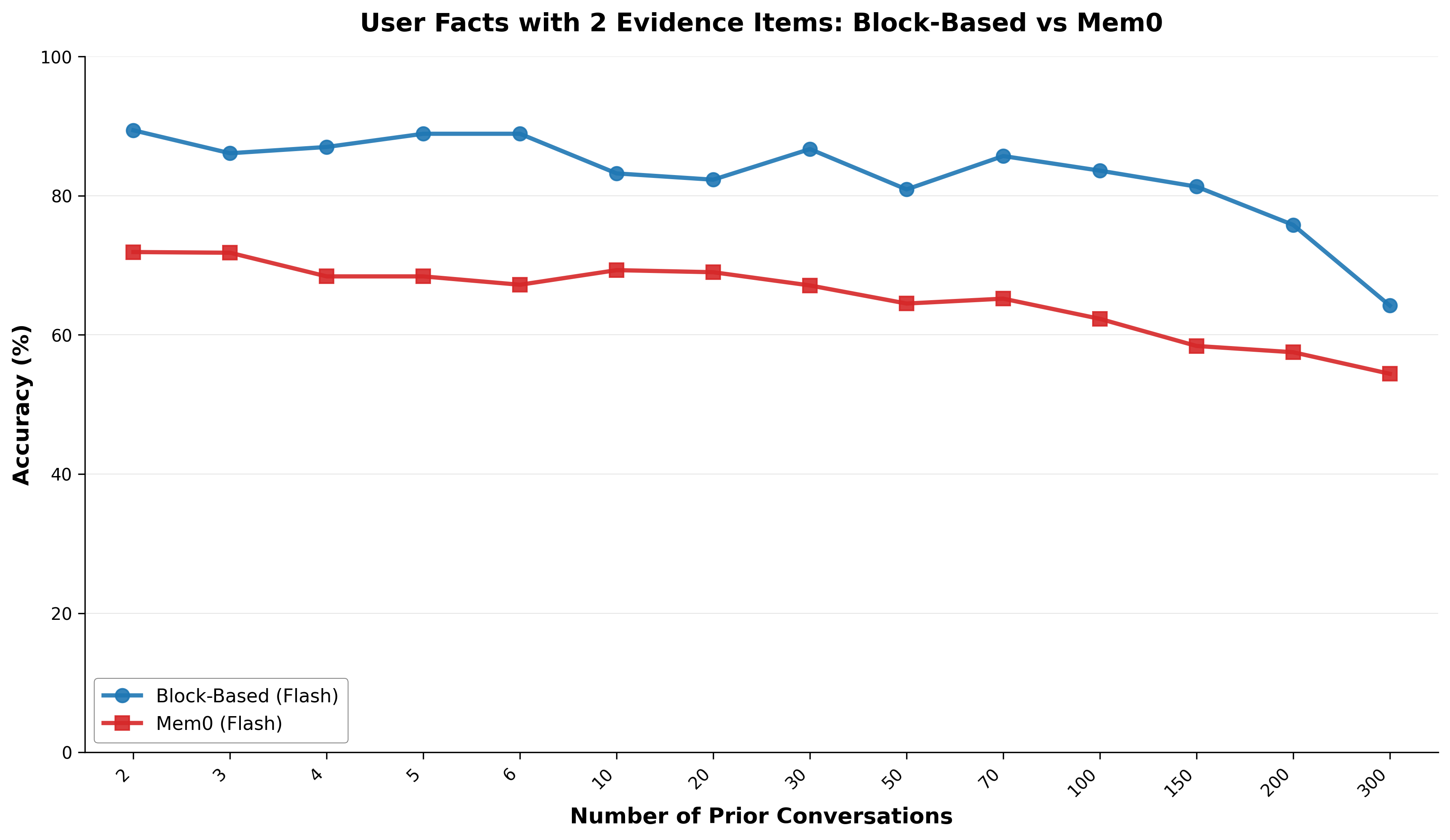}
\caption{Block-Based vs Mem0 (User Facts, 2 Evidence Items)}
\label{fig:block_vs_mem0}
\end{figure}

Despite reduced performance relative to long context on complex tasks, block-based extraction
still significantly outperforms Mem0, achieving 64.2\% accuracy versus Mem0's 54.4\% at 300
conversations---a 9.8 percentage point advantage that demonstrates the superiority of hybrid
approaches over pure RAG systems.

\textbf{Latency Reduction Through Parallelization:}

A critical advantage of block-based extraction is dramatic latency reduction through parallelization.
Long context memory suffers from high latency as history length increases---processing 300 conversations
sequentially creates severe bottlenecks, with response times exceeding 80 seconds for Pro models.
Block-based extraction eliminates this bottleneck by dividing the 300 conversations into 30 blocks
of 10 conversations each, enabling parallel processing across all blocks simultaneously. Since each
block processes only 10 conversations instead of 300, individual extraction completes in seconds
rather than minutes. With sufficient parallel compute resources, the entire extraction phase completes
in the time it takes to process a single block---a potential 30x latency reduction. This parallelization
transforms conversational memory from a sequential bottleneck into a scalable, production-ready system
suitable for real-time applications.

\paragraph{Single-Pass Extraction Strategy}

Single-pass extraction processes all conversations as one block, extracts all relevant
information in a single pass, then generates the answer from the extracted context.

\begin{figure}[h]
\centering
\includegraphics[width=0.8\textwidth]{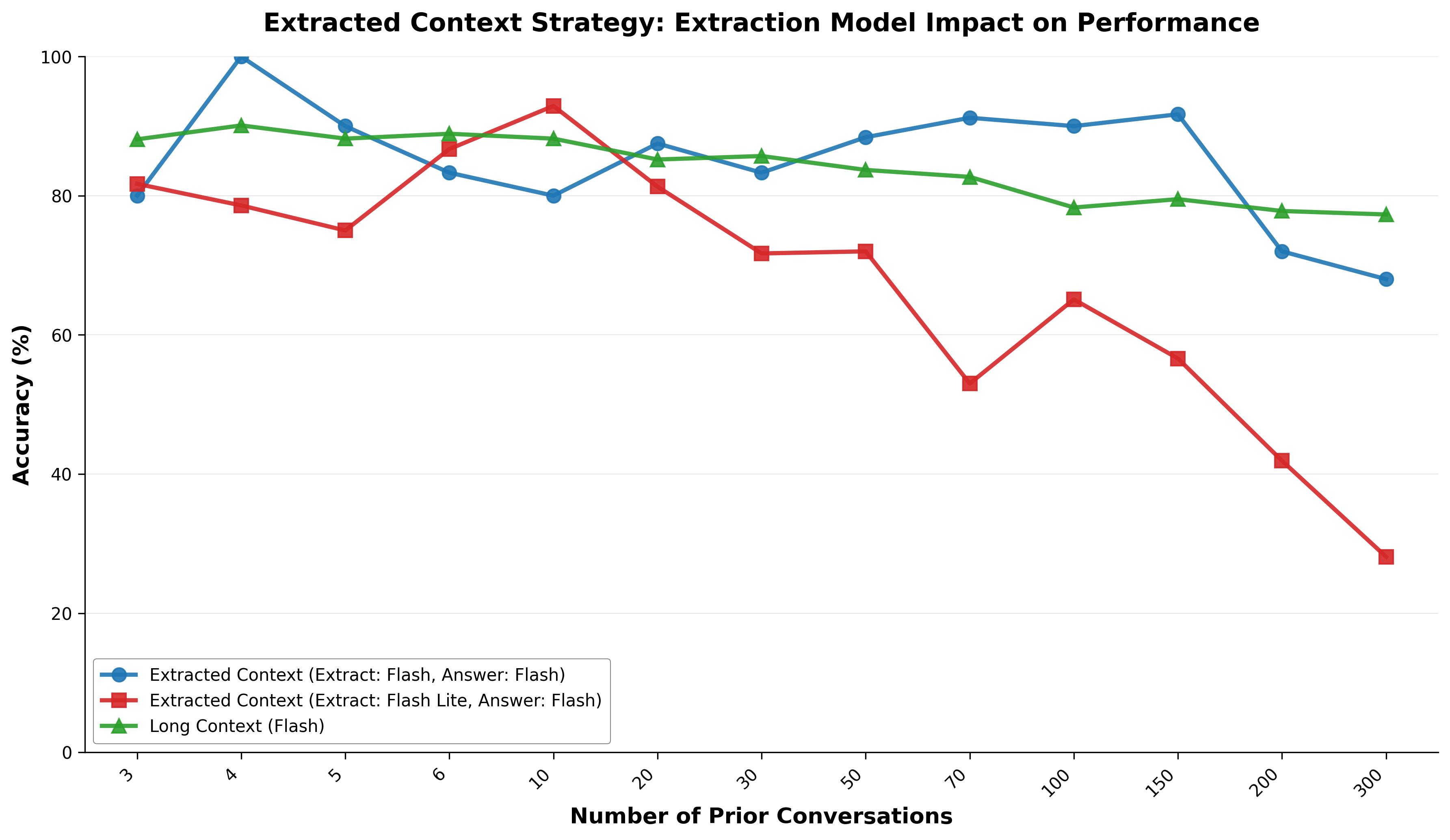}
\caption{Single-Pass Extraction Performance (Preferences, 1 Evidence Item)}
\label{fig:single_pass_preferences}
\end{figure}

The single-pass extraction strategy, evaluated on preference tasks with 1 evidence item, shows
different patterns compared to block-based extraction on implicit connections:

\begin{enumerate}
\item \textbf{Extracted Context (Extract: Flash, Answer: Flash)} - Blue line: Using Flash for both
   extraction and answering achieves 68.0\% accuracy at 300 conversations, only 9.3 percentage
   points below the long context baseline, demonstrating reasonable performance for preference tasks.

\item \textbf{Extracted Context (Extract: Flash Lite, Answer: Flash)} - Red line: Using Flash Lite for
   extraction severely degrades performance to 28.1\% accuracy, a 49.2 percentage point drop from
   the long context baseline (77.3\%). This confirms that extraction model capacity is critical
   for maintaining accuracy.

\item \textbf{Long Context (Flash)} - Green line: The baseline achieves 77.3\% accuracy, outperforming
   both extracted context configurations for this preference task.
\end{enumerate}

The 39.9 percentage point performance gap between Flash and Flash Lite extraction aligns with
our earlier model size analysis, where Flash Lite showed 24-31 percentage point degradation on
memory tasks. This consistent pattern confirms that Flash Lite lacks the minimum capacity required
for reliable information extraction---whether processing full conversations directly or extracting
relevant context. The extraction phase inherits the same model capacity requirements we identified
for long context systems, reinforcing that mid-tier models like Flash represent the practical
minimum for conversational memory tasks.

\textbf{Optimizing for High-Stakes Applications with Pro Answering:}
\begin{figure}[h]
\centering
\includegraphics[width=0.8\textwidth]{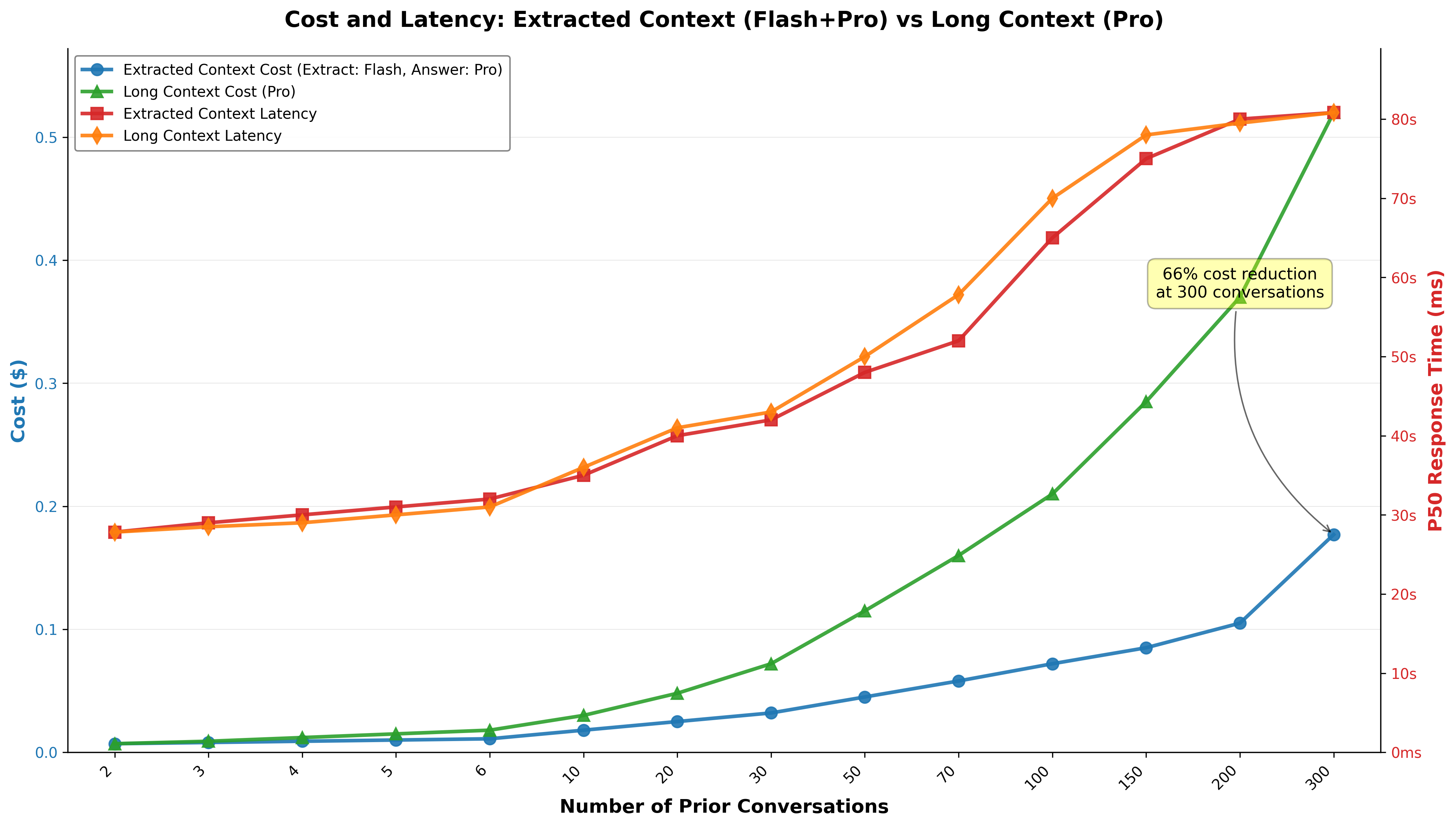}
\caption{Cost and Latency: Extracted Context vs Long Context Pro}
\label{fig:pro_cost_latency}
\end{figure}

For applications requiring the highest accuracy, our two-phase architecture enables a cost-optimized
configuration: using Flash for extraction and Pro for answering. This hybrid approach achieves 58.6\%
accuracy on complex implicit connection tasks with 2 evidence items---within 6.1 percentage points of
full Pro long context (64.7\%)---while reducing costs by 66\% at 300 conversations (\$0.177 vs \$0.520).
Latency remains comparable ($\sim$80s) as both approaches must process similar token volumes, but the
dramatic cost reduction makes Pro-level reasoning economically viable for production deployments.
This configuration demonstrates how the two-phase architecture enables flexible optimization: affordable
extraction with Flash processes the bulk of tokens, while Pro focuses its superior reasoning capacity
on the distilled context for final answer generation.

\textbf{Key advantages of hybrid approaches:}
\begin{itemize}
\item \textbf{Dramatic latency reduction:} Block-based extraction enables parallel processing of conversation
  chunks, potentially achieving 30x latency reduction compared to sequential long context processing---
  transforming 80+ second response times into sub-3 second responses with parallel compute
\item \textbf{Cost optimization with trade-offs:} Block-based extraction with Flash Lite reduces per-token
  costs by approximately 10x compared to Flash, but introduces additional fixed request overhead
  from processing 30 blocks instead of one context. The net cost benefit depends on conversation
  volume---at 300 conversations, token savings typically outweigh request overhead, but shorter
  histories may see increased total costs due to multiple API calls
\item \textbf{Answer generation efficiency:} Both strategies reduce the answering context by 10-13x, processing
  only extracted relevant information instead of full conversation history. This dramatic reduction
  frees valuable context window budget for other critical components---detailed user profiles, complex
  system instructions, relevant documentation, or multiple examples---that would be impossible to include
  when conversation history alone consumes most of the context limit. For production systems needing
  rich context beyond just conversation history, this 10-13x reduction transforms infeasible deployments
  into practical solutions
\item \textbf{Model flexibility:} The two-phase architecture enables using smaller, cheaper models for
  extraction when task characteristics allow, while preserving high-quality models for the
  critical answer generation phase
\item \textbf{Scalability:} Extraction cost remains relatively constant regardless of conversation
  history length, unlike long context's linear growth
\item \textbf{Performance optimization:} Block-based extraction with appropriate models can actually
  outperform long context (70.8\% vs 63.5\%), suggesting that structured extraction helps
  models focus on relevant information
\end{itemize}

\textbf{Performance preservation:} Block-based extraction maintains accuracy comparable to
full long context (70-75\% at 300 conversations) while dramatically reducing
computational requirements.

\subsubsection{Observed Transition Patterns}
Our analysis suggests approximate ranges where different memory approaches tend to make
sense:
\begin{itemize}
\item \textbf{First $\sim$30 conversations:} Long context generally performs best---good accuracy with
  reasonable cost/latency
\item \textbf{Around 30-150 conversations:} Long context still works but gets pricey---block-based
  extraction starts looking attractive for cost-conscious applications
\item \textbf{Roughly 150-300 conversations:} Things get interesting here---block-based extraction
  often strikes a nice balance, though pure RAG becomes tempting if cost really matters
\item \textbf{Beyond $\sim$300 conversations:} RAG-based systems usually become the practical choice
  despite their accuracy trade-offs
\end{itemize}

\subsection{Discussion}
\label{subsec:discussion}

The 35-40 percentage point accuracy gap between naive long context and sophisticated RAG
systems reveals a fundamental misalignment between how memory systems are built and how
they should evolve. The field has optimized for eventual scale while ignoring that most
users will never exceed 150 conversations---essentially building highways before local roads.
Our data shows clear architectural transition points ($\sim$30, $\sim$150, $\sim$300 conversations) that
align with natural usage patterns: the first month of daily interaction, the first year
of regular use, and power users. These boundaries should drive system design, not
theoretical scalability limits. Critically, mid-tier models (Flash-class) achieve within
2-6\% of premium model accuracy at 3.7x lower cost, suggesting the field is also overspending
on model capacity when the bottleneck is retrieval architecture, not reasoning capability.

The economic implications challenge conventional wisdom about memory system deployment.
While RAG achieves 95x cost reduction at scale, this comes with accuracy dropping to 30-45\%---
essentially making the system wrong more often than right on preference and implicit
reasoning tasks. For enterprise applications in healthcare, finance, or customer service,
the cost of incorrect responses far exceeds the infrastructure savings. Our hybrid extraction
architectures offer a middle path: maintaining 70-75\% accuracy while enabling 30x latency
reduction through parallelization, transforming 80-second response times into sub-3-second
interactions. This suggests a dual-track deployment strategy where recent conversations
use high-accuracy long context while historical data employs cost-effective RAG, rather
than forcing all interactions through the same pipeline.

The small-corpus characteristic of conversational memory enables optimizations that violate
RAG best practices but deliver superior results. Exhaustive search, complete reranking,
and full attention mechanisms---all considered anti-patterns in document retrieval---become
not just feasible but optimal for conversation histories. This asymmetry extends to
development practices: memory systems should be tested and optimized on realistic
conversation volumes (10-150 interactions) rather than stress-tested at scales most users
never reach. The field's focus on billion-token benchmarks for systems that start from
zero tokens reflects a fundamental misunderstanding of the problem space.

Looking forward, our results suggest memory research should pivot from architectural
complexity to progressive evolution. Instead of proposing novel retrieval mechanisms
that claim theoretical advantages, the field needs threshold-aware systems that
automatically transition between approaches as conversation volume grows. The surprising
effectiveness of naive policies isn't an anomaly to be fixed but a feature to be
exploited---most memory problems are small-data problems masquerading as big-data
challenges. The path forward isn't through more sophisticated RAG but through
recognizing that conversational memory, for most of its lifecycle, doesn't need RAG at all.

\section{Conclusion}
\label{sec:conclusion}

We present a comprehensive benchmark with 75,336 question-answer pairs for evaluating
conversational memory systems and demonstrate empirically that memory systems are
fundamentally RAG systems operating on progressively growing, initially small corpora.
This unique characteristic enables an important but underappreciated finding: naive long
context approaches can more than double the accuracy of sophisticated RAG-based memory
systems when operating on conversation histories under 150 interactions.

Our analysis suggests general patterns for memory system design: long context approaches
tend to work best for the initial 30-150 conversations or so, hybrid block-based
extraction can often extend effectiveness to around 300 conversations while preserving
decent accuracy, and RAG-based approaches typically become the practical choice beyond
that despite their accuracy penalties. Additionally, we show that medium-tier language
models provide equivalent memory performance to premium models at dramatically lower
costs, suggesting that memory applications can optimize costs without sacrificing quality.

These findings challenge the prevailing wisdom that sophisticated retrieval mechanisms
are necessary for conversational memory and highlight an underexplored opportunity in
exploiting the unique characteristics of small, growing corpora rather than simply
applying general RAG solutions. For the significant portion of real-world usage that falls within
150 conversations, simplicity not only suffices but excels. The framework's extensibility
and complete reproducibility ensure it can evolve alongside advancing AI capabilities
while providing quantitative insights for optimizing memory systems across different
cost-latency-accuracy requirements.

\section{Data and Code Availability}

The ConvoMem benchmark dataset containing 75,336 question-answer pairs is publicly available at:
\url{https://huggingface.co/datasets/Salesforce/ConvoMem}

The complete evaluation framework and code for reproducing all experiments is available at:
\url{https://github.com/SalesforceAIResearch/ConvoMem}

\bibliographystyle{plain}
\bibliography{references}

\appendix

\section{Technical Implementation Details}

\subsection{Coherent Synthetic Data Generation Pipeline}

Our synthetic data generation pipeline implements a carefully orchestrated three-phase
process designed to balance diversity with quality control. This section provides the
complete technical implementation details for researchers seeking to replicate or extend
our benchmark.

\subsubsection{Persona Generation}
The pipeline begins with creating detailed synthetic personas that span enterprise roles:
IT administrators, financial analysts, customer service representatives, project managers,
and sales executives. These personas include industry-specific knowledge, professional
goals, and organizational contexts that serve as the foundation for all subsequent content.
This enterprise focus ensures the benchmark captures the complexity of business communication
patterns rather than just casual consumer interactions.

\subsubsection{Phase 1: Use Case Generation}
We request batches of scenarios---typically 50-100 per persona---in a single LLM call. This
batching strategy is crucial for diversity: the model can see all previously generated
scenarios within that batch and ensure each new scenario explores different aspects of
the persona's life, avoiding repetition and maximizing coverage. Each scenario includes
a natural question the user would ask, ensuring diversity across the entire dataset from
the outset.

\subsubsection{Phase 2: Evidence Core Generation}
This phase deliberately shifts to individual processing. We transform one use case at a
time into a refined question-answer pair along with evidence messages that must all come
from the same speaker (either User or Assistant depending on evidence type). This isolation
allows the model to focus deeply on the specific scenario without distraction, ensures
thorough validation of each generated core, and enables precise retry logic when validation
fails. Built-in checks ensure multi-message evidence is non-redundant and that questions
become unanswerable if any piece is removed.

\subsubsection{Phase 3: Conversation Generation}
The final phase embeds evidence messages naturally within full 80-120 message conversations,
maintaining strict requirements:
\begin{itemize}
\item The Nth evidence message must appear in the Nth conversation
\item Messages are copied exactly to preserve integrity
\item Evidence-containing and filler conversations use identical generation methods
\end{itemize}

This separation of concerns---first defining what information needs testing, then generating
the evidence core, and finally embedding it in natural dialogue---ensures both quality and
consistency while enabling parallel generation at each phase for efficiency.

\subsection{Comprehensive Validation Framework}

Our validation framework operates at three critical junctures in the generation pipeline,
with less than 5\% of generated examples passing all stages.

\subsubsection{Structural Validation (Phase 2)}
During Evidence Core Generation, we validate:
\begin{itemize}
\item Correct number of evidence messages
\item Proper speaker assignments (User vs Assistant)
\item Non-redundancy in multi-message evidence
\item Dependency requirements (removing any piece makes question unanswerable)
\end{itemize}

Any core failing these checks triggers immediate regeneration, preventing propagation of
malformed data through the pipeline.

\subsubsection{Conversation Embedding Validation (Phase 3)}
During Conversation Generation, we verify evidence messages appear exactly once in their
designated conversations using:
\begin{itemize}
\item Exact string matching as the primary method
\item Fuzzy matching with edit distance thresholds for minor variations
\item Partial matching strategies for handling formatting differences
\item Strict ordering enforcement (Nth evidence in Nth conversation)
\end{itemize}

\subsubsection{Verification System}
The final validation employs multiple independent checks executed by different language
models. The standard verification suite:

\begin{enumerate}
\item \textbf{Positive validation}: Models must answer questions correctly when given only the
   relevant evidence conversations in a small context window
\item \textbf{Negative validation}: Models cannot answer without access to those conversations
\item \textbf{Consistency requirement}: At least two consecutive correct answers from each model
   (eliminates lucky guesses)
\end{enumerate}

\subsubsection{Category-Specific Validation Approaches}

\textbf{Multi-message evidence}: Implements exhaustive partial evidence validation, systematically
removing each conversation to confirm all pieces are essential. If removing any single
conversation still allows correct answers, the entire evidence item is rejected.

\textbf{Changing evidence}: Verifies that the latest conversation containing the final update
is strictly necessary---removing it should make the question unanswerable with outdated
information. We also validate that intermediate evidence messages actually address the
question topic rather than being random filler.

\textbf{Abstention categories}: Inverts the typical validation logic, recognizing ``I don't know''
responses as correct while treating any specific answer as failure.

\textbf{Preference and implicit connections}: Employs rubric-based evaluation where a judge
model assesses response quality rather than requiring exact matches. This allows for
evaluating nuanced responses that capture the spirit of preferences without requiring
specific wording.

\subsubsection{Model Selection for Validation}
We deliberately use small, cost-effective models for validation (GPT-4o-mini, Gemini Flash)
rather than premium models. This ensures that even basic models can answer our questions
with 100\% accuracy when given the evidence directly, guaranteeing that any performance
degradation in actual testing reflects memory system capabilities rather than model
intelligence limitations. During generation, we rotate through multiple such models with
each required to answer correctly multiple times consecutively.

\subsection{Framework Architecture and Extensibility}

\subsubsection{Modular Design}
Each evidence category is fully encapsulated within a single class containing:
\begin{itemize}
\item Generation prompts specific to that category
\item Validation strategy and rules
\item Category-specific parameters and thresholds
\end{itemize}

The underlying framework remains universal across all scenarios, handling:
\begin{itemize}
\item Persona management and storage
\item Conversation generation orchestration
\item Evaluation infrastructure and metrics
\item Parallel execution and retry logic
\item Statistical tracking and reporting
\item Cost monitoring and optimization
\end{itemize}

\subsubsection{Adding New Categories}
New memory challenges can be implemented without touching the core codebase. For example,
adding implicit connections required only:
\begin{enumerate}
\item Writing prompts that describe the desired behavior
\item Implementing a validation strategy using rubric-based evaluation
\item Defining category-specific parameters
\end{enumerate}

The framework automatically handles parallel generation, retry logic, statistical tracking,
and output formatting.

\subsubsection{Cost and Performance Characteristics}
\begin{itemize}
\item \textbf{Generation time}: Each new category typically requires 3 hours of generation time
\item \textbf{Dataset regeneration}: Complete benchmark regeneration requires approximately 24 hours
\item \textbf{Cost tracking}: Comprehensive monitoring for both generation and evaluation phases
\item \textbf{Parallelization}: 10 threads per persona for generation, 40 threads for evaluation
\item \textbf{Data format}: All data exports to standardized JSON format for reproducibility
\end{itemize}

\subsubsection{Integration Capabilities}
The framework's flexible data model enables:
\begin{itemize}
\item Variable conversation history lengths (2-300 conversations)
\item Easy integration of existing benchmarks (successfully converted LongMemEval and LoCoMo)
\item Support for different memory system architectures (long context, RAG, hybrid)
\item Configurable evaluation parameters and early stopping criteria
\end{itemize}

This extensibility ensures the benchmark can evolve alongside advancing AI capabilities
while maintaining backward compatibility and reproducibility.

\subsection{Benchmark Execution Infrastructure}

\subsubsection{Batch Execution and Statistical Balancing}
Since evaluations often terminate early when statistical significance is reached or budgets
are exhausted, the framework's sophisticated batch execution system divides test cases into
balanced batches that ensure equal representation across all context sizes (from 2 to 300
conversations). Without this balancing, smaller contexts would be evaluated first and might
exhaust the budget before larger contexts could be tested, creating statistical skew. By
processing short-context (under 30 conversations) and long-context test cases in parallel
thread pools, the system achieves faster convergence to representative results while
maintaining load balance across computational resources.

\subsubsection{Intelligent Early Termination}
The framework implements intelligent early termination through multiple statistical
conditions that monitor spending, accuracy patterns, and convergence indicators. Rather
than blindly executing all test cases, the system continuously evaluates whether sufficient
statistical confidence has been achieved, checking conditions such as: monotonic decrease
in accuracy as context size increases (indicating consistent memory degradation patterns),
minimum sample sizes per context (at least 50 correct answers per context size), and
progressive cost thresholds that balance statistical significance with evaluation expense.
This adaptive approach typically reduces evaluation costs by 40-60\% while maintaining
statistical validity.

\subsubsection{Cost Tracking and Monitoring}
The framework tracks detailed cost metrics throughout execution, ultimately providing
comprehensive cost-accuracy profiles that reveal the economic trade-offs of different
memory architectures. Real-time cost monitoring includes input tokens, output tokens,
cached tokens (with cache hit ratios often exceeding 90\% for repeated context patterns),
and dollar costs computed using current model pricing, providing immediate feedback on
evaluation expenses and enabling budget-conscious research.

\subsubsection{Performance Instrumentation}
The framework's comprehensive instrumentation enables detailed performance analysis
across different memory architectures. For instance, tail latencies for long context
approaches can exceed 30 seconds at 300 conversations, making them impractical for
interactive applications despite acceptable median performance. Cache utilization
tracking showed that conversation prefixes achieve 85-95\% cache hit rates in providers
like Gemini, dramatically reducing costs for evaluations that test multiple questions
against the same conversation history.

\subsubsection{Multi-Level Caching Infrastructure}
The evaluation framework's caching infrastructure operates at multiple levels to enable
efficient iterative research. Test case caching allows researchers to modify evaluation
parameters---such as switching memory systems or models---without regenerating the expensive
test case dataset, reducing setup time from hours to seconds. The caching system maintains
versioned test case sets, automatically invalidating caches when evidence generators or
mixing strategies change, ensuring reproducibility while maximizing reuse.

A particularly valuable capability is the framework's re-judging functionality, which
decouples expensive memory system execution from answer evaluation. Since memory systems
take the longest time to generate responses (especially for large contexts), the framework
preserves these responses in structured logs that can be re-evaluated with different
judging criteria, models, or prompts without re-running the memory system. This enables
rapid iteration on evaluation criteria---researchers can experiment with stricter or more
lenient judging, test different judge models, or debug evaluation prompts in minutes
rather than hours.

\subsubsection{Robustness and Recovery}
During execution, the framework implements intelligent retry logic with exponential backoff
for transient failures, distinguishing between retryable errors (rate limits, temporary
network issues) and permanent failures (invalid prompts, out-of-context errors). This
robustness is critical when running evaluations that might span several hours and encounter
various failure modes. The framework also provides periodic checkpoint exports every 5
minutes, enabling evaluation recovery from interruptions without losing progress, and
generating cumulative statistics that can be analyzed even from partial runs.

\subsubsection{Comprehensive Logging and Analysis}
The framework's comprehensive logging and analysis capabilities extend beyond simple
accuracy metrics. Each evaluation run generates structured logs that capture the complete
decision-making process: which conversations were retrieved, what information was extracted,
how the final answer was synthesized, and why incorrect answers occurred. This forensic
capability proved invaluable for understanding failure modes---revealing, for instance, that
changing facts questions fail not because systems can't find updates but because they
struggle to determine which update is most recent when multiple changes occur across a
conversation history.

The framework exports results in multiple formats optimized for different analyses: CSV
files for statistical analysis with columns for all metrics per context size, JSON logs
preserving complete question-answer pairs for qualitative analysis, and aggregated
summaries showing trends across evidence categories. This comprehensive execution framework
transforms memory evaluation from an expensive, opaque process into a systematic,
observable, and economically viable research methodology.

\section{Detailed Memory-RAG Convergence Analysis}

\subsection{Temporal Reasoning}

Both conversational memory and document RAG systems struggle with temporal reasoning---the
fundamental challenge of correctly placing information onto a timeline to ensure its
validity and relevance. Both domains face identical temporal challenges that manifest regardless of
corpus type. In conversational systems, memory must resolve ambiguous temporal
references like ``yesterday,'' ``next week,'' or ``since our last chat'' relative to
conversation timestamps, while tracking evolving user states where preferences change and
old information becomes invalid---what researchers term ``historical noise.'' Document RAG
systems face precisely the same challenges: queries containing vague temporal expressions
like ``recently'' or ``after the war'' require contextual interpretation, while the phenomenon
of ``knowledge drift'' means yesterday's facts may be false today.

This shared temporal challenge has driven both fields through remarkably parallel
evolutionary trajectories in solution development. Both began with simple recency
heuristics---assuming the newest information is most relevant---before independently
discovering this approach's brittleness when queries require understanding event order
or duration. Both fields have independently evolved toward explicit timeline construction:
conversational AI developed ``time-aware memorization through timeline summarization'' where
the TReMu framework extracts events from dialogue sessions and associates them with inferred
dates to form a retrievable temporal memory, while RAG systems evolved ``Temporal Agents''
that process documents to extract fact triplets annotated with valid\_at and invalid\_at
timestamps, building queryable timelines of facts.

\subsection{Implicit Information Extraction}

Both conversational memory and RAG systems face the fundamental challenge of resolving
semantically incomplete inputs where critical information must be inferred from external
context. In conversational memory, this manifests as cross-turn dependencies where
current utterances require inference from dialogue history---what recent surveys formalize
as the challenge of ``dialogue coherence and context maintenance across multiple turns''
requiring both conversation-level and turn-level memory. In RAG, complex queries require
traversing implicit entity chains across documents.

Advanced dialogue state tracking has evolved beyond simple slot-filling to address
semantic incompleteness through sophisticated reasoning mechanisms. The PREMem framework
shifts complex reasoning from inference to memory construction, extracting fine-grained
memory fragments and establishing explicit relationships (extensions, transformations,
implications) before storage. Parallel developments in RAG show remarkable architectural
similarity. The MIND framework implements memory stores to ensure ``retrieved entities remain
accessible across reasoning steps,'' using memory-aware filtering with Chain-of-Thought
validation and confidence-based ranking.

\subsection{Knowledge Update Mechanisms}

Both conversational memory and RAG systems face an identical fundamental challenge:
managing information that becomes outdated or contradicted over time. In conversational
systems, users routinely correct previous statements (``Actually, the meeting is at 3 PM,
not 2 PM''), while RAG systems must navigate document versioning where newer publications
supersede older ones. Bae et al. formalized this as requiring explicit REPLACE and DELETE
operations for conversational memory, directly paralleling how RAG systems must handle
document updates.

The architectural convergence extends beyond problem similarity to nearly identical
solution patterns. Both domains independently evolved from simple recency heuristics
(``newest information wins'') to sophisticated mechanisms involving temporal metadata,
source provenance, and explicit conflict resolution. This parallel evolution suggests
that managing evolving information requires fundamentally similar architectural patterns
regardless of whether the knowledge comes from user conversations or document
collections.

\subsection{Graph-Structured Representations}

Both conversational memory and document RAG systems have independently evolved
remarkably similar graph-based architectures. In conversational memory, systems like
Zep construct ``temporally-aware dynamic knowledge graphs'' with entities and
relationships extracted from dialogue episodes, while Mem0 employs a two-stage pipeline---an
``Entity Extractor Module'' followed by a ``Relationship Generator Component''---to transform
conversations into structured triplets like (Alice, lives\_in, San\_Francisco). Nearly
identical architectures appear in document RAG: Microsoft's GraphRAG framework uses
LLMs to parse text chunks and ``identify all entities'' and ``all relationships among the
identified entities.''

\section{Architectural Analysis of Memory-RAG Convergence}

This appendix provides detailed analysis of how conversational memory and RAG systems have converged on nearly identical architectures despite operating on fundamentally different scales. We examine four key architectural patterns that have emerged independently in both domains.

\subsection{Temporal Reasoning Mechanisms}

Both memory and RAG systems face the challenge of resolving ambiguous temporal references. In conversations, ``yesterday'' or ``last week'' require maintaining precise timelines. In document collections, ``recently announced'' or ``upcoming quarter'' demand similar temporal grounding. Both domains have evolved from simple timestamp comparison to sophisticated timeline construction with explicit temporal entity extraction and resolution.

\subsection{Implicit Information Extraction}

The challenge of extracting information from incomplete or indirect statements has driven both fields toward increasingly sophisticated reasoning frameworks. Memory systems must infer that ``I work at TechCorp'' and ``TechCorp launched an AI product'' means the user works on AI products. Similarly, RAG systems must connect scattered document mentions to construct complete knowledge graphs.

\subsection{Knowledge Update Mechanisms}

Managing contradictions and corrections has led both domains to implement explicit versioning systems. Memory systems track when preferences change (``I used to love sushi but developed an allergy''), while RAG systems handle document updates and corrections. Both have moved beyond simple recency heuristics to implement REPLACE and DELETE operations with provenance tracking.

\subsection{Graph-Structured Representations}

The independent evolution toward knowledge graphs in both domains---Zep and Mem0 for conversations, GraphRAG for documents---reflects fundamental requirements for capturing entity relationships. These systems use nearly identical two-stage pipelines: entity extraction followed by relationship generation, producing structured representations that enable complex reasoning.

\section{Multi-Message Evidence Examples}

This appendix provides comprehensive examples of multi-message evidence patterns across all benchmark categories, demonstrating how information distributed across multiple conversation turns requires sophisticated memory integration.

\subsection{User Facts - Multi-Message Examples}

\textbf{Example 1 (2 messages):}
\begin{itemize}
\item Evidence 1: ``I just started a new job at DataCorp as a senior engineer''
\item Evidence 2: ``The commute from my place in Brooklyn takes about 45 minutes''
\item Question: ``How long is my commute to my new job?''
\item Expected: ``Your commute to DataCorp takes about 45 minutes''
\end{itemize}

\textbf{Example 2 (3 messages):}
\begin{itemize}
\item Evidence 1: ``My daughter Emma just turned 5 last month''
\item Evidence 2: ``We're planning to enroll Emma in kindergarten this fall''
\item Evidence 3: ``The local elementary school has a great kindergarten program''
\item Question: ``Where is my daughter starting school?''
\item Expected: ``Emma will be starting kindergarten at the local elementary school''
\end{itemize}

\subsection{Assistant Facts - Multi-Message Examples}

\textbf{Example 1 (2 messages):}
\begin{itemize}
\item Evidence 1: ``Python 3.12 introduced improved error messages with more helpful suggestions''
\item Evidence 2: ``The new error messages in Python 3.12 highlight the exact location of syntax errors''
\item Question: ``What debugging improvements were made in Python 3.12?''
\item Expected: ``Python 3.12 introduced improved error messages that highlight exact error locations and provide helpful suggestions''
\end{itemize}

\textbf{Example 2 (3 messages):}
\begin{itemize}
\item Evidence 1: ``The Mediterranean diet emphasizes olive oil, vegetables, and whole grains''
\item Evidence 2: ``Studies show the Mediterranean diet reduces heart disease risk by 30\%''
\item Evidence 3: ``The diet also includes moderate amounts of fish and poultry''
\item Question: ``What are the key components and benefits of the Mediterranean diet?''
\item Expected: ``The Mediterranean diet emphasizes olive oil, vegetables, whole grains, with moderate fish and poultry, and reduces heart disease risk by 30\%''
\end{itemize}

\subsection{Changing Facts - Multi-Message Examples}

\textbf{Example 1 (3 messages):}
\begin{itemize}
\item Evidence 1: ``The product launch is scheduled for March 15th''
\item Evidence 2: ``Due to supply chain issues, we're moving the launch to April 1st''
\item Evidence 3: ``Good news - we resolved the issues and can launch on March 20th''
\item Question: ``When is the product launching?''
\item Expected: ``March 20th''
\end{itemize}

\textbf{Example 2 (4 messages):}
\begin{itemize}
\item Evidence 1: ``Our budget for Q3 is \$2.5 million''
\item Evidence 2: ``Finance approved an increase to \$3 million for Q3''
\item Evidence 3: ``Due to market conditions, Q3 budget is reduced to \$2.8 million''
\item Evidence 4: ``Final Q3 budget confirmed at \$2.9 million after negotiations''
\item Question: ``What's our Q3 budget?''
\item Expected: ``\$2.9 million''
\end{itemize}

\subsection{Abstention - Multi-Message Examples}

\textbf{Example 1 (2 messages):}
\begin{itemize}
\item Evidence 1: ``I'm considering three vendors for our CRM system''
\item Evidence 2: ``Each vendor has different pricing models and feature sets''
\item Question: ``Which CRM vendor did we select?''
\item Expected: ``You haven't selected a CRM vendor yet - you're still considering three options''
\end{itemize}

\textbf{Example 2 (3 messages):}
\begin{itemize}
\item Evidence 1: ``The team is researching cloud migration strategies''
\item Evidence 2: ``We're evaluating AWS, Azure, and GCP''
\item Evidence 3: ``Cost analysis for each platform is still pending''
\item Question: ``What's the total cost of our cloud migration?''
\item Expected: ``The cloud migration cost hasn't been determined - cost analysis is still pending''
\end{itemize}

\subsection{Preferences - Multi-Message Examples}

\textbf{Example 1 (2 messages):}
\begin{itemize}
\item Evidence 1: ``I prefer morning meetings before 10 AM when I'm most alert''
\item Evidence 2: ``Please avoid scheduling me for meetings on Friday afternoons''
\item Question: ``When should I schedule your team review?''
\item Answer (Rubric): ``Should suggest morning before 10 AM, Monday-Thursday. Must not suggest Friday afternoon''
\end{itemize}

\textbf{Example 2 (2 messages):}
\begin{itemize}
\item Evidence 1: ``I'm vegetarian and don't eat any meat or fish''
\item Evidence 2: ``I also have a severe nut allergy''
\item Question: ``What restaurant should we go to for the team lunch?''
\item Answer (Rubric): ``Must suggest vegetarian-friendly options. Must not suggest seafood/steakhouses or places known for nut-heavy dishes''
\end{itemize}

\subsection{Implicit Connections - Multi-Message Examples}

\textbf{Example 1 (2 messages):}
\begin{itemize}
\item Evidence 1: ``I live in Seattle and love coffee''
\item Evidence 2: ``The weather here is perfect for staying indoors with a warm drink''
\item Question: ``What's my favorite drink in my city?''
\item Expected: ``Your favorite drink in Seattle is coffee''
\end{itemize}

\textbf{Example 2 (3 messages):}
\begin{itemize}
\item Evidence 1: ``My wife is pregnant and due in 3 months''
\item Evidence 2: ``We just moved to a new city and don't know anyone yet''
\item Evidence 3: ``I've been working 60-hour weeks on this critical project''
\item Question: ``Any recommendations for our anniversary celebration next month?''
\item Answer (Rubric): ``Should suggest low-key, local options. Must not suggest travel, adventure activities, or alcohol-focused venues''
\end{itemize}

\textbf{Example 3 (3 messages, implicit):}
\begin{itemize}
\item Evidence 1: ``I mentioned last month I'm training for the Boston Marathon''
\item Evidence 2: ``My knee has been bothering me during long runs''
\item Evidence 3: ``The race is only 6 weeks away''
\item Question: ``Should I increase my training intensity?''
\item Answer (Rubric): ``Should recommend caution due to knee issues and proximity to race. Must not suggest aggressive intensity increases''
\end{itemize}

\end{document}